\documentclass[11pt,letterpaper]{article}
\usepackage[in]{fullpage}

\usepackage{Packages/usual_packages}
\usepackage{Packages/usual_macros}

\usepackage[numbers]{natbib}

\usepackage[utf8]{inputenc} 
\usepackage[T1]{fontenc}    
\usepackage{hyperref}       
\usepackage{url}            
\usepackage{booktabs}       
\usepackage{amsfonts}       
\usepackage{nicefrac}       
\usepackage{microtype}      
\usepackage{xcolor}         
\usepackage{thmtools}
\usepackage{thm-restate}

\title{Implicit Bias of SGD for Diagonal Linear Networks: \\  a Provable Benefit of Stochasticity}

\author{%
\vspace{-0.5em}
\\
  Scott Pesme \\
  EPFL\\
  \texttt{scott.pesme@epfl.ch}
  \and 
  \vspace{-0.5em}
  \\
  \qquad 
Loucas Pillaud-Vivien \\
\qquad  EPFL\\
\qquad  \texttt{loucas.pillaud-viven@epfl.ch}
\and \\
  Nicolas Flammarion \\
  EPFL\\
  \texttt{nicolas.flammarion@epfl.ch} \\
}
\date{}

\begin{document}

\maketitle

\begin{abstract}

Understanding the implicit bias of training algorithms is of crucial importance in order to explain the success of overparametrised neural networks. 
In this paper, we study the dynamics of stochastic gradient descent over diagonal linear networks through its continuous time version, namely stochastic gradient flow. 
We explicitly characterise the solution chosen by the stochastic flow and prove that it always enjoys better generalisation properties than that of gradient flow.
Quite surprisingly, we show that the convergence speed of the training loss controls the magnitude of the biasing effect: the slower the convergence, the better the bias. 
To fully complete our analysis, we provide convergence guarantees for the dynamics. 
We also give experimental results which support our theoretical claims. 
Our findings highlight the fact that structured noise can induce better generalisation and 
they help explain the greater performances of stochastic gradient  descent over gradient descent observed in practice.

\end{abstract}


\section{Introduction}

%

Understanding the performance of neural networks 
 is certainly one of the
most thrilling challenges for the current machine learning community. 
From the theoretical point of view, progress has been made in several directions:
we have a better 
functional analysis description of neural networks~\cite{bach2017curse} and we steadily understand the convergence of training algorithms~\cite{mei2018mean,chizat2018global} as well 
as the role of initialisation~\cite{jacot2018ntk,chizat2019lazy}. Yet there remain many unanswered questions. 
One of which is why do the currently used training algorithms converge to solutions which generalise well,
 and this with very little use 
of explicit regularisation \citep{recht2017understanding}. 


To understand this phenomenon, the concept of \emph{implicit bias} has emerged: 
if over-fitting is benign, it must be because the optimisation procedure converges towards some particular global minimum which enjoys good generalisation 
properties. 
Though no explicit regularisation is added, the algorithm is implicitly selecting a particular solution: this is
referred to as the implicit bias of the training procedure. 
The implicit regularisation of several algorithms has been studied, the simplest and most emblematic being that of gradient descent and stochastic gradient 
descent in the least-squares 
framework: they both converge towards the global solution which has the lowest squared distance from the initialisation. 
For logistic regression on separable data, Soudry et al. show in the seminal paper \citep{soudry2018implicit} that gradient descent selects the max-margin classifier.
This type of result has then been extended to neural networks and to other frameworks. Overall, characterising the implicit bias 
of gradient methods has almost always come down to unveiling mirror-descent like structures which underlie  the algorithms. 


While mostly all of the results focus on gradient descent, it must be pointed out that this full batch algorithm is not used in practice for neural networks since it does not lead to solutions which generalise well~\citep{keskar2017large}.
Instead, results on stochastic gradient descent, which is widely used and shows impressive results, are still missing or unsatisfactory.
%
This has certainly to do with the fact that grasping the nature of the noise induced by the stochasticity of the algorithm is 
particularly hard: 
it mixes properties from the model's architecture, the data's distribution and the loss. 
In our work, by focusing on simplified neural networks,  we answer to the following fundamental questions: do SGD's and GD's implicit bias differ? 
What is the role of SGD's noise over the algorithm's implicit bias? 

The simplified neural networks which we consider are  diagonal linear neural networks; despite their simplicity 
they have become popular since they
already enable to grasp the complexity of more general networks.
Indeed, they highlight important aspects of the theoretical concerns of modern machine learning: 
the neural tangent kernel regime, the roles of over-parametrisation, of the initialisation and of the step size.
For a regression problem where we assume the existence of an interpolating solution,
we study stochastic gradient descent  through its continuous version, namely stochastic gradient flow (SGF). 
Though the continuous modelling of SGD has not yet led to many fruitful results compared to the well studied gradient flow, 
we believe it is because capturing the essence of the stochastic noise is particularly difficult. 
It has generally been done in a non realistic and over simplified manner, such as considering constant and isotropic noise.
In our work, we attach peculiar attention to the adequate modelling of the noise.  Tools from Itô calculus are then leveraged in order 
to derive exact formulas, quantitative bounds and interesting interpretations for our problem.

\subsection{Main contributions and paper organisation.} 

\begin{figure}[t]
\centering
\begin{minipage}[c]{.4\linewidth}
\hspace*{-10pt}
\includegraphics[width=\linewidth]{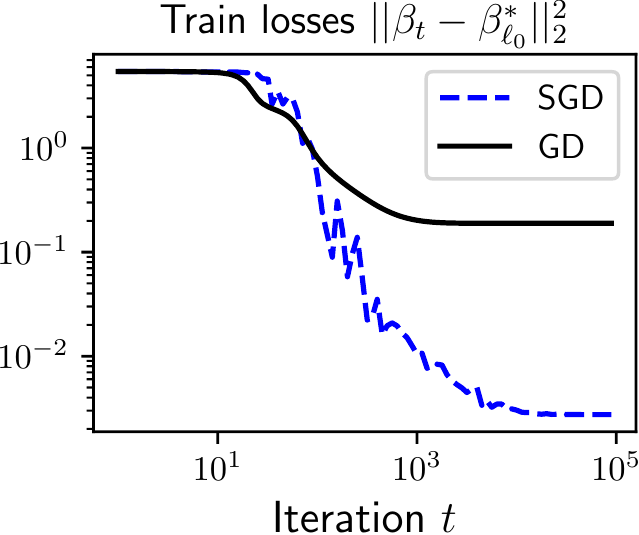}
   \end{minipage}
   \hspace*{30pt}
   \begin{minipage}[c]{.4\linewidth}
\includegraphics[width=\linewidth]{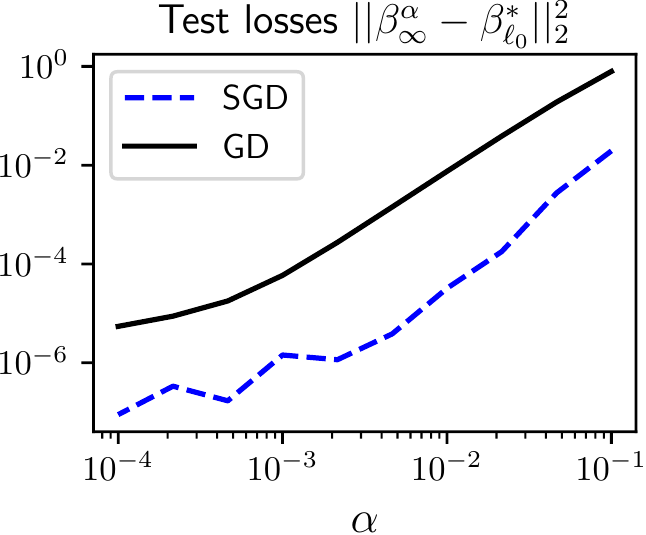}
   \end{minipage}
  \caption{Sparse regression with $n = 40$, $d = 100$, $\Vert \beta^*_{\ell_0} \Vert_0 = 5$, $x_i \sim \mathcal{N}(0, I)$ 
  $ y_i = x_i^{\top } \beta^*_{\ell_0}  $ . \textit{Left}: for initialisation scale $\alpha = 0.05$, SGD converges towards a solution which generalises better than GD.  
  \textit{Right}: for different values of the initialisation scale $\alpha$, 
  the solution recovered by SGD has better validation loss than that of GD. The sparsifying effect due to their implicit 
  biases differ by more than an order of magnitude. See \Cref{subsection:experim_setup} for the precise experimental setup.
    }
     \label{fig:pink_flamingo}
\end{figure}

In Section~\ref{sec:setup}, we start by introducing the setup of our problem as well 
as the continuous modelisation of stochastic gradient descent. Then, in Section~\ref{sec:implicit_bias}, 
we state our main result on the implicit bias of the stochastic gradient flow. 
We informally formulate it here and illustrate it in Figure~\ref{fig:pink_flamingo}:
\begin{thm}[Informal]
Stochastic gradient flow over diagonal linear networks converges with high probability to a zero-loss solution 
which enjoys better generalisation properties than the one obtained by gradient flow. 
Furthermore, the speed of convergence of the training loss controls the magnitude of the biasing effect:
the slower the convergence, the better the bias.
\end{thm}
\setcounter{thm}{0}
%
%
Unlike previous works \citep{gunasekar2018characterizing,woodworth2020kernel}, in addition to characterising the implicit bias effect of SGF,  
we also prove the convergence of the iterates towards a zero-loss solution with high-probability. To accomplish this, we leverage in Section~\ref{sec:dynamical_properties}
the fact that the iterates follow a stochastic continuous mirror descent with a time-varying potential. 
We support our results experimentally and validate our model in Section~\ref{sec:going_further}.
%
%

\subsection{Related work}

As recalled, implicit bias has a recent history that has been initiated by the seminal work \cite{soudry2018implicit} 
on max-margin classification with $\log$-loss for a linear setup and separable data.
This work has been extended to other architectures, \emph{e.g.} multiplicative parametrisations \cite{gunasekar2018characterizing}, 
linear networks \cite{ji2018gradient} and more general homogeneous neural networks \cite{Lyu2020Gradient,chizat2020implicit}. 
In \cite{woodworth2020kernel} the authors show that the scale of the initialisation leads to an interpolation 
between the neural tangent kernel regime \citep{jacot2018ntk,chizat2019lazy} (which is a linear regression on fixed features) 
leading to $\ell_2$ minimum norm solutions and the rich regimes leading to $\ell_1$ minimum norm solutions. 
Note that these works focus
on full batch gradient descent (or flow) and are deeply linked to mirror descent. 


While the links between SGD's stochasticity and generalisation have been looked into in numerous works~\citep{mandt2016stochastic,jastrzebski2018three,he2019control,hoffer2017train,kleinberg2018alternative}, no such explicit characterisation of implicit regularisation have ever been given. It has been empirically observed that SGD often outputs models which generalise better than GD~\citep{keskar2017large,jastrzebski2018three,he2019control}. One suggested explanation is that SGD is prone to pick flatter solutions than GD and that bad generalisation solutions are correlated with sharp minima, i.e., with strong curvature, while good generalisation solutions are correlated with flat minima, i.e., with low curvature~\citep{hochreiter1997flat,keskar2017large}. This idea has been further investigated by adopting a random walk on random landscape modelling~\citep{hoffer2017train},  by suggesting that SGD's noise is smoothing the loss landscape, thus eliminating the sharp minima~\citep{kleinberg2018alternative}, by considering a dynamical stability perspective~\citep{wu2018sgd} or by interpreting SGD as a diffusion process~\citep{he2019control,jastrzebski2018three,chaudhari2018stochastic}. Recently, label-noise has been shown to influence the implicit bias of SGD, by biasing the solution towards the origin for quadratically-parameterized models~\citep{haochen2020shape} or by  implicitly regularising the expected squared norm of the gradient of the model with respect to the weights~\citep{blanc2020implicit}. Thus, if the notion of implicit bias of GD is fairly well understood both in the cases of regression and classification, it remains unclear for SGD, and its explicit characterisation is missing. 

The linear diagonal neural networks we consider have been studied in the case of gradient descent~\cite{vavskevivcius2019implicit} and stochastic gradient descent with label noise~\cite{haochen2020shape}. In both cases the authors show that this model has the ability to implicitly bias the training procedure to help retrieve a sparse predictor. The link between gradient descent and mirror descent for this model has been initiated by \cite{ghai2020exponentiated} and further exploited by the same author in \cite{NEURIPS2020_e9470886,NEURIPS2020_024d2d69} for its sparse inducing property. 

Contrary to the deterministic case, the modelling of stochastic gradient descent as a stochastic differential equation is quite recent, see~\cite{mandt2016stochastic,jastrzebski2018three}. However, as highlighted by 
\citep{ali2020implicit}, early attempts often suffer from the drawback that they model the noise using a constant covariance matrix. On the contrary, state dependant noise has now become the legitimate manner for modelling SGD as a stochastic gradient flow and it is shown in \cite{JMLR:v20:17-526} that it can be done consistently. 
Yet, noise modelling still remains the principal issue \citep{wojtowytsch2021stochastic} as it influences largely the behaviour of the dynamics~\cite{chaudhari2018stochastic,cheng2020stochastic}.

\vspace{-0.6em}

\subsection{Notations}

For input data $(x_1, \dots, x_n) \in (\R^{d})^n$ and output $(y_1, \dots, y_n) \in \R^n$, 
we denote respectively $ X \in \R^{n \times d}$ the design matrix whose $i$-th row is feature $x_i \in \R^d$ and $y \in \R^n$ 
the vector of outputs. 
$\R_+^*$ denotes the set of strictly positive real numbers.
For $p = {1,2} $, the $\ell_p$-norm of $x \in \R^d$ is $\|x\|^p_p = \sum_i^d |x_i|^p$. 
The operations $\odot$ will stand for coordinate-wise product between vector: $[u \odot v ]_i = u_i v_i$ and 
$u^2 = u \odot u$.
For $p \in \N^*$, we also define $u^p := u \odot \hdots \odot u$, the $p$ times product of $u$ with itself. 
All inequalities between vectors should be understood value by value.
For $f,g \in \R$, the existence of $C>0$ such that $f\leq C g$ and  $Cg \leq f$ will be denoted $f \leq O(g)$ and $\Omega(g) \leq f$ respectively. We shall use the symbole $\widetilde{O}$ when this is true up to $\log$ factors.
For a vector $u \in \mathbb{R}^d$, $\diag(u)$ denotes the $d \times d$
diagonal matrix which has its diagonal equal to $u$. For 
a matrix $M \in \R^{d \times d}$, $\diag(M)$ denotes the vector $(M_{1 1}, \dots, M_{d d}) \in \R^d$.
The indexed vector $\beta^*$ will stand for any $\beta$ interpolating the data, i.e. any vector in the  affine space
$\{\beta \in \R^d\ s.t,\ X\beta = Y \}$ of dimension at least $d - n$. Out of all these, let 
$\betasparse = \underset{\beta \in \R^d \ \text{s.t.} \ X \beta = y}{ \argmin} \Vert \beta \Vert_1$. 
For $z$ any vector, $z_\infty$ or $z^\infty$ will always designate of $\underset{t \to \infty}{\lim}\ z_t$.

\vspace{-0.7em}


\section{Setup and preliminaries}


\label{sec:setup}

\subsection{Architecture and algorithm.}



\paragraph*{Overparametrised noiseless regression.} We consider a linear regression problem with outputs $(y_1, \dots , y_n) \in \R^n$ and inputs $(x_1, \dots, x_n) \in (\R^{d})^n$.
%
%
%
%
%
We study an overparametrised setting ($ n < d$)
and assume that there exists at least one 
interpolating parameter  $\beta^* \in \R^d$ which perfectly fits the training set, i.e.  $y_i = \langle \beta^*, x_i \rangle$ for all $1 \leq i \leq n$.
%
 We parametrise the regression vector $\beta$ as $\beta_w$ with $w \in \R^p$. We will see that though in the end  
our final models $x \mapsto \langle \beta_w, x \rangle$
are classical linear models whatever the parametrisation $w \mapsto \beta_w$, the choice of this parametrisation has crucial consequences 
on the solution recovered by the learning algorithms. We study the quadratic loss and the overall loss is written as:
\begin{align*}
    L(w) = L(\beta_w) := \frac{1}{4 n} \sum_{i = 1}^n (\langle \beta_w, x_i \rangle- y_i)^2 = \frac{1}{4 n} \sum_{i = 1}^n \langle \beta_w - \beta^*, x_i \rangle^2, 
\end{align*}
where by abuse of notation we use $L(w) = L(\beta_w)$. 
%
\paragraph*{$2$-layer diagonal linear network.} 
The simplest parametrisation of $\beta_w$ is to consider $\beta_w = w$ which corresponds to the classical least-squares framework. 
It is well known 
that in this case, many first order methods (GD, SGD, with and without momentum)
will converge towards the same
solution: we say that they have the same implicit bias. 
This is experimentally not the case for neural networks where SGD has been shown to lead to solutions which have better generalisation properties
compared to GD~\citep{keskar2017large}.
To theoretically confirm this observation, we study a simple non-linear parametrisation: $\beta_w = w_+^2 - w_-^2$ with $w = [w_+, w_-]^\top \in \R^{2 d}$.
We point out that it is 
 $2$-positive homogeneous and that it is equivalent to the parametrisation $\beta_{u, v} = u \odot v$ with $u, v \in \R^d$.
 It should 
be thought of a simplified linear network of depth $2$ (see ~\citep[Section~$4$]{woodworth2020kernel} for more details). 
We consider two weight vectors $w_+$ and $w_-$ (and not only $\beta_w = w^2$) in order 
to ensure that our final linear predictor parameter $\beta_w$ can take negative values.
%
For the sake of completeness, the study of diagonal linear networks of arbitrary depth $p\geq 3$ is done  in \Cref{app:subsec:depth_p}. 
Also note that additionally to being a toy neural model, it has received recent attention 
for its practical ability to induce sparsity~\cite{vavskevivcius2019implicit,NEURIPS2020_024d2d69,haochen2020shape} 
or to solve phase retrieval problems~\cite{NEURIPS2020_e9470886}.


\paragraph*{Stochastic Gradient Descent.} With this quadratic parametrisation, the loss now rewrites as: $L(w) = \frac{1}{4 n} \sum_{i = 1}^n \langle w_+^2 - w_-^2 - \beta^*, x_i \rangle^2.$ 
Note that despite its simplicity, this loss is non convex and its minimisation is non trivial.
The algorithm we shall consider is the well known SGD algorithm, where for a step size $\gamma > 0$:
\begin{equation}
\label{eq:SGD}
 \left.
    \begin{array}{ll}
        &w_{t+1, +} = w_{t, +} - \gamma  \langle \beta_w - \beta^*, x_{i_t} \rangle \  x_{i_t} \odot w_{t, +}  \\
        &w_{t+1, -} = w_{t, -} + \gamma  \langle \beta_w - \beta^*, x_{i_t} \rangle \  x_{i_t} \odot w_{t, -} 
    \end{array}
 \right. \qquad \text{where  } i_t \sim \mathrm{Unif}(1, n).
\end{equation}
It is convenient to rewrite this recursion as
\begin{align}
\label{eq:SGD_rewritten}
    w_{t+1, \pm} = w_{t, \pm} - \gamma \nabla_{w_\pm} L(w_t) \pm \gamma \diag(w_{t, \pm}) X^{\top} \xi_{i_t}(\beta_{t}),
\end{align}
where $ \xi_{i_t}(\beta) = - \big (
 \langle \beta - \beta^*, x_{i_t} \rangle  \mathbf{e}_{i_t} - \mathbb{E}_{i_t} \big [ \langle \beta - \beta^*, x_{i_t} \rangle \mathbf{e}_{i_t}  \big ] \big ) \in \R^{n}$ 
  is a zero-mean \emph{multiplicative} noise
which vanishes at any global optimum ($\mathbf{e}_{i}$ denotes the $i^{\text{th}}$ element of the canonical basis).
We point out that all the results we shall give hold for any initialisation such that $w_{t=0, +} = w_{t=0, -} \in \R^d$, under which we have that  
$\beta_{w_{t = 0}} = 0$. 
To understand
under what conditions the SGD procedure converges  
and towards which point it does, 
we shall consider its continuous counterpart which has the advantage of leading to clean 
and intuitive calculations. We highlight the fact that we consider 
a bath-size equal to $1$  for clarity, however all our analysis holds for mini-batch SGD (with and without replacement)
simply by considering an effective step-size $\gamma_\mathrm{eff}$ instead of $\gamma$, 
this is clearly explained in \Cref{app:subsec:details_SDE_model}.





\subsection{Stochastic gradient flow}
\label{subsec:SDE_model}


Continuous time modelling of sequential processes offer a large set of tools, such as derivation, 
which come in helpful to understand the dynamics of the processes.
This has led to a large part of the recent literature to consider 
continuous gradient flow in order and understand the behaviour of gradient descent on complicated architectures such as 
neural nets.  
However, the continuous time modelling of stochastic gradient descent is more challenging: it requires to add on top of the gradient flow a diffusion term whose covariance matches the one of SGD. Hence, it is fundamental to understand its structure and scale.


%
%
\paragraph*{Understanding the noise's structure.}
As seen in equation \eqref{eq:SGD_rewritten}, evaluated at $w_{\pm}$,  the stochastic noise $\gamma \diag(w_{\pm}) X^{\top} \xi_{i_t}(w)$ has two 
main characteristics which we want to preserve:
\begin{itemize}
  \item It belongs to $\mathrm{span}(w_{\pm} \odot x_1, \dots, w_{\pm} \odot x_n )$
  \item It has covariance $\Sigma_{_{\textrm{SGD}}}(w_{\pm}) := \gamma^2 \diag(w_\pm) X^{\top}  \text{Cov}_{i_t}(\xi_{i_t}(\beta))  X \diag(w_\pm) \in \R^{d \times d}$ 
\end{itemize}
It remains to understand the structure of the covariance of $\xi_{i_t}$ which has the following closed form:
$\text{Cov}_{i_t}(\xi_{i_t}(\beta)) 
= \frac{1}{n} \diag( \langle \beta - \beta^* , x_i \rangle^2)_{1 \leq i \leq n} -  \frac{1}{n^2} \big ( \langle \beta - \beta^* , x_i \rangle \langle \beta - \beta^* , x_j \rangle \big )_{1 \leq i, j \leq n}  $. 
We identify the two key facts:
(i) it is diagonal at the leading $n^{-1}$ order  and 
(ii) its trace is linked to the loss as $\mathrm{Var}_{i_t}( \Vert \xi_{i_t}(\beta) \Vert_2 ) = \frac{4}{n} L(\beta) + O(\frac{1}{n^2})$.
This leads us in modelling $\xi_{i_t}(\beta)$'s covariance matrix as $\frac{4}{n} L(\beta) I_n$ as it preserves
 these two characteristics \footnote{the general case is discussed in \Cref{app:subsection:E:general_SDE_model}}. Finally this brings us to consider the following 
 modelling of the overall noise's structure: $\Sigma_{_{\textrm{SGD}}}(w_{\pm}) \cong \frac{4}{n} \gamma^2 L(w) [\diag(w_\pm) X^{\top}]^{\otimes 2}$.

\paragraph*{Stochastic differentiable equation modelling.}
Guided by the previous considerations, we study the following stochastic gradient flow: 
\begin{equation}
  \label{eq:SGF}
  \begin{aligned}
  \dd w_{t,+} &= - \nabla_{w_+} L(w_t) \diff t + 2 \sqrt{\gamma n^{-1} L (w_t)}\  w_{t,+} \odot [X^\top \dd  B_t]  \\
  \dd w_{t,-} &= - \nabla_{w_-} L(w_t) \diff t - 2 \sqrt{\gamma n^{-1} L (w_t)}\  w_{t,-} \odot [X^\top \dd  B_t],
  \end{aligned}
  \end{equation}
where $\diff B_t$ is a standard $\R^n$ Brownian motion. 
The SDE is a perturbed gradient flow with a diffusion term that is defined such that
 its Euler discretisation with step size 
$\gamma$ leads to a Markov Chain whose covariance 
exactly matches SGD's noise covariance $\Sigma_{_{\textrm{SGD}}}(w_{\pm})$.
We refer to~\cite{JMLR:v20:17-526} or~\citep{kloeden1992stochastic} for the technical details regarding consistency of such a procedure in the limit of small step sizes. This stochastic differential equation is the starting point of the analysis.

\section{The implicit bias of the stochastic gradient flow}
\label{sec:implicit_bias}


\paragraph*{Implicit bias and hyperbolic entropy.} 
To understand the relevance of the main result and how stochasticity induces a preferable bias, we start by recalling some 
known results for gradient flow. 
In \cite{woodworth2020kernel} it is shown, assuming global convergence, that the solution selected by the 
gradient flow initialised at $\alpha \in \R^d$ and denoted $\beta_\infty^\alpha$ solves a constrained optimisation problem 
involving the \emph{hyperbolic entropy} introduced by \cite{ghai2020exponentiated}:  
\begin{align}
\label{eq:potential_function}
    \beta_\infty^\alpha = \underset{\beta \in \R^d \ \text{s.t.} \ X\beta = y}{\argmin} \phi_\alpha (\beta) := \frac{1}{4} \big[  \sum_{i = 1}^d \beta_i \arcsinh(\frac{\beta_i}{2 \alpha_i^2}) - \sqrt{\beta_i^2 + 4 \alpha_i^4} \big ] ,
\end{align}
Though the hyperbolic entropy function has a non-trivial expression, its principal characteristic is that it interpolates between the $\ell_1$ and the $\ell_2$ norms according to the scale of $\alpha$. 
More precisely for 
$\alpha \in \R$~\footnote{If $\alpha \in \R$ we consider the abuse of notation 
$\phi_\alpha :=\phi_{\alpha \mathbf{1}}$.}: $\phi_\alpha(\beta) \underset{\alpha \to 0}{\sim} \frac{1}{2} \ln \big( \frac{1}{\alpha} \big ) \Vert \beta \Vert_1$ and
$\phi_\alpha(\beta) \underset{\alpha \to + \infty}{=} 2 \alpha^2 + \frac{1}{4 \alpha^2}  \Vert \beta \Vert_2^2 + o(\alpha^{-2})$.
We refer to \cite[Theorem 2]{woodworth2020kernel} for more details on the asymptotic analysis. 
The implicit optimisation problem \eqref{eq:potential_function} therefore highlights 
the fact that the initialisation scale of the weights controls the shape of the recovered solution.
Small initialisations lead to low $\ell_1$-norm solutions which are known to induce good generalisation properties: this is 
what is often referred to as the \emph{rich regime}.
Large initialisations lead to low $\ell_2$-norm solutions: this is referred to as the \emph{kernel regime} or 
\emph{lazy regime} 
in which the weights move only very slightly. The dynamics of the gradient flow 
are then very similar to the one of kernel linear regression with the kernel depending on the initialisation 
\cite{jacot2018ntk,chizat2019lazy}. 
Overall, to retrieve a sparse solution, one should initialise with the smallest $\alpha$ possible. 
However, as is clearly explained in \cite{woodworth2020kernel}, 
it is important to stress out that there is a generalisation / optimisation tradeoff:
the point $w = 0$ happens to be a saddle point for the loss and a smaller $\alpha$ will lead to a longer training time.


\paragraph*{Main result.} In the main theorem we show that, for an initialisation scale $\alpha$,  
the stochasticity of SGF biases the flow towards solutions 
which still minimise the hyperbolic entropy. 
However, what is remarkable is that it does so with an 
effective parameter $\alpha_{\infty}$ which is strictly smaller than $\alpha$. 
The recovered solution therefore minimises an optimisation problem which has better sparsity inducing properties than that of gradient flow.

\vspace{1em}

\begin{figure}[ht]
\centering
\begin{minipage}[c]{.32\linewidth}
\hspace*{-15pt}
\includegraphics[trim={0 0 0 0}, width=\linewidth]{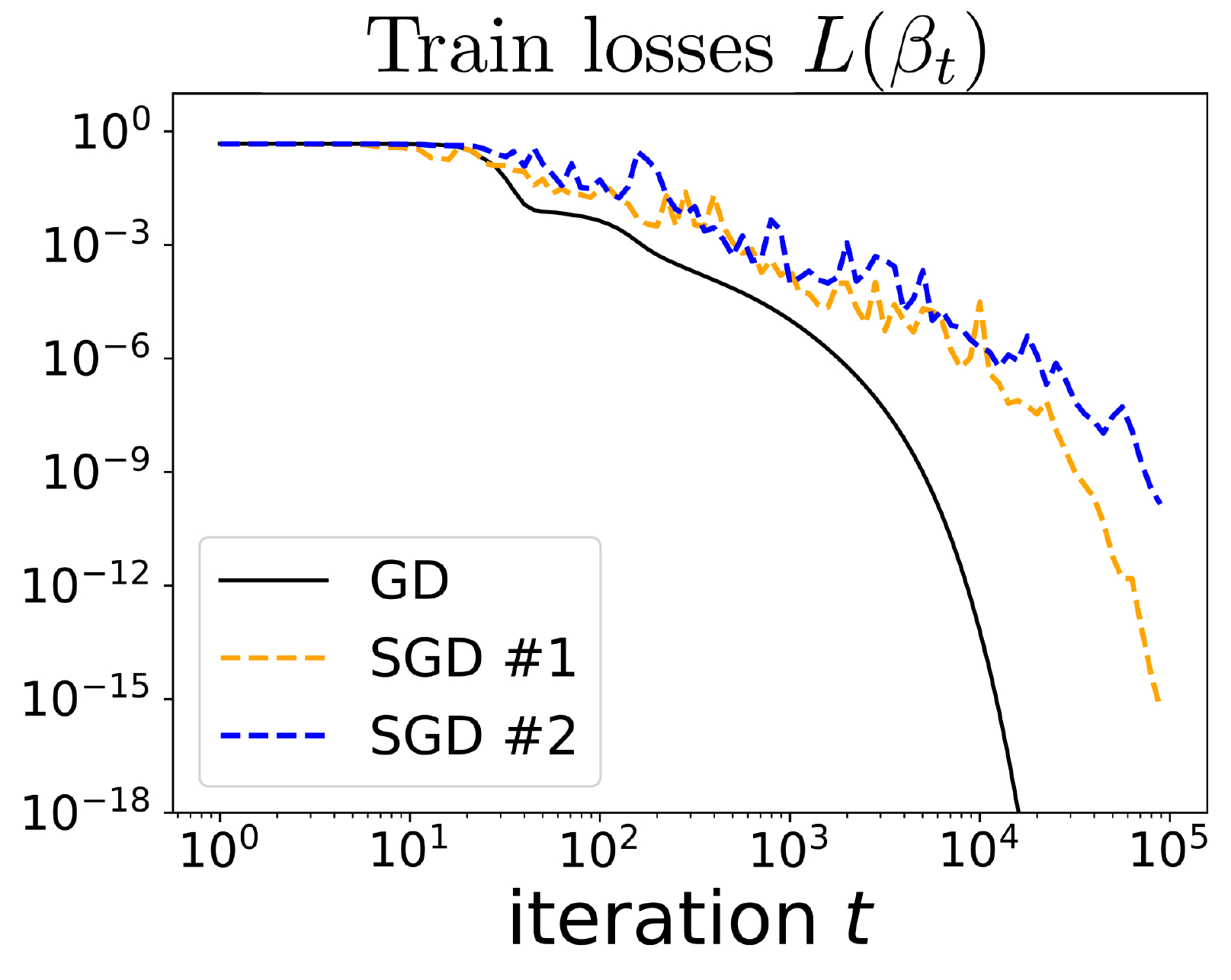}
   \end{minipage}
   \hspace*{-15pt}
   \begin{minipage}[c]{.32\linewidth}
    \vspace*{-6.5pt}
\includegraphics[width=0.913\linewidth]{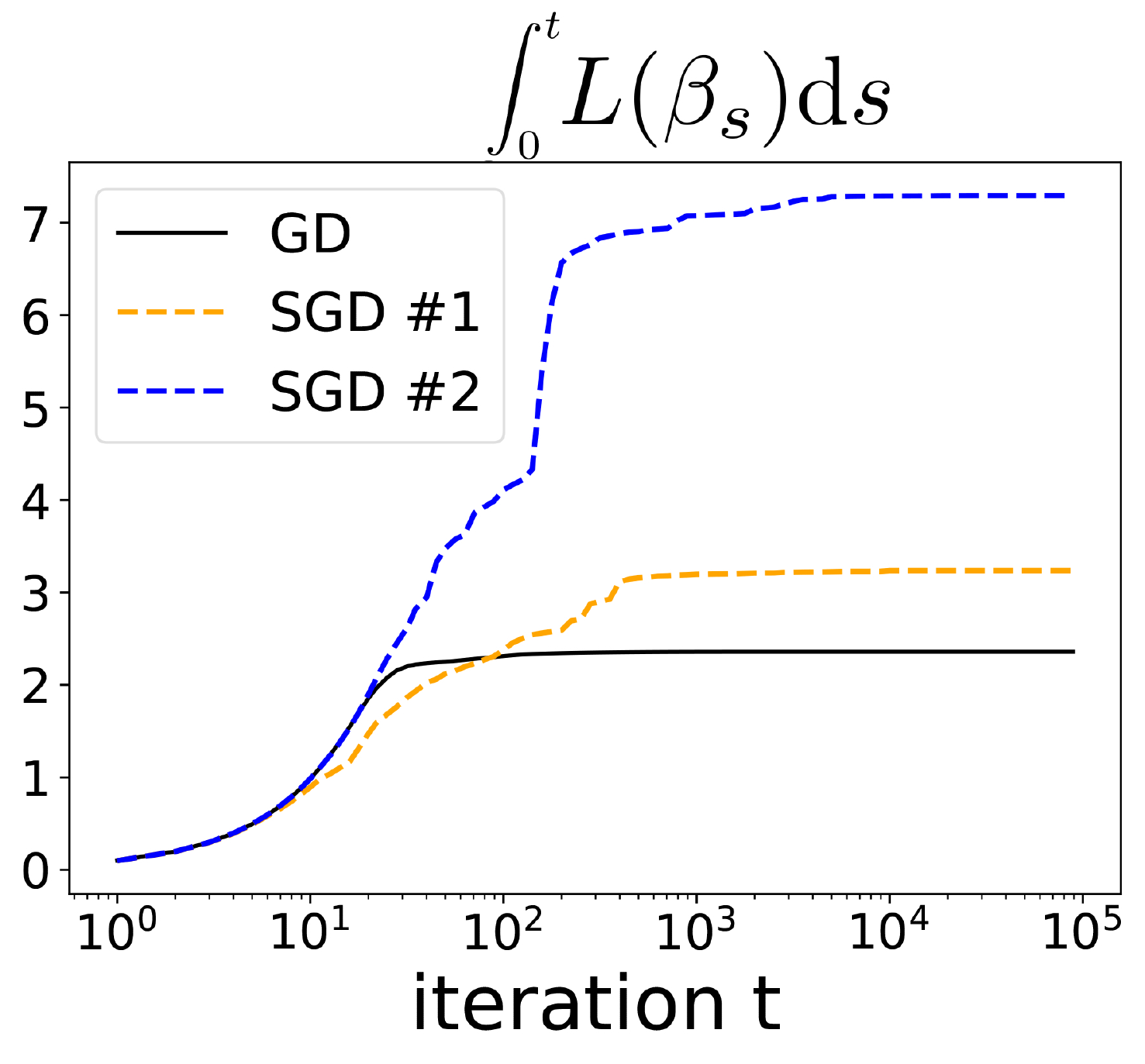}
   \end{minipage}
    \hspace*{-15pt}
   \begin{minipage}[c]{.32\linewidth}
\includegraphics[width=\linewidth]{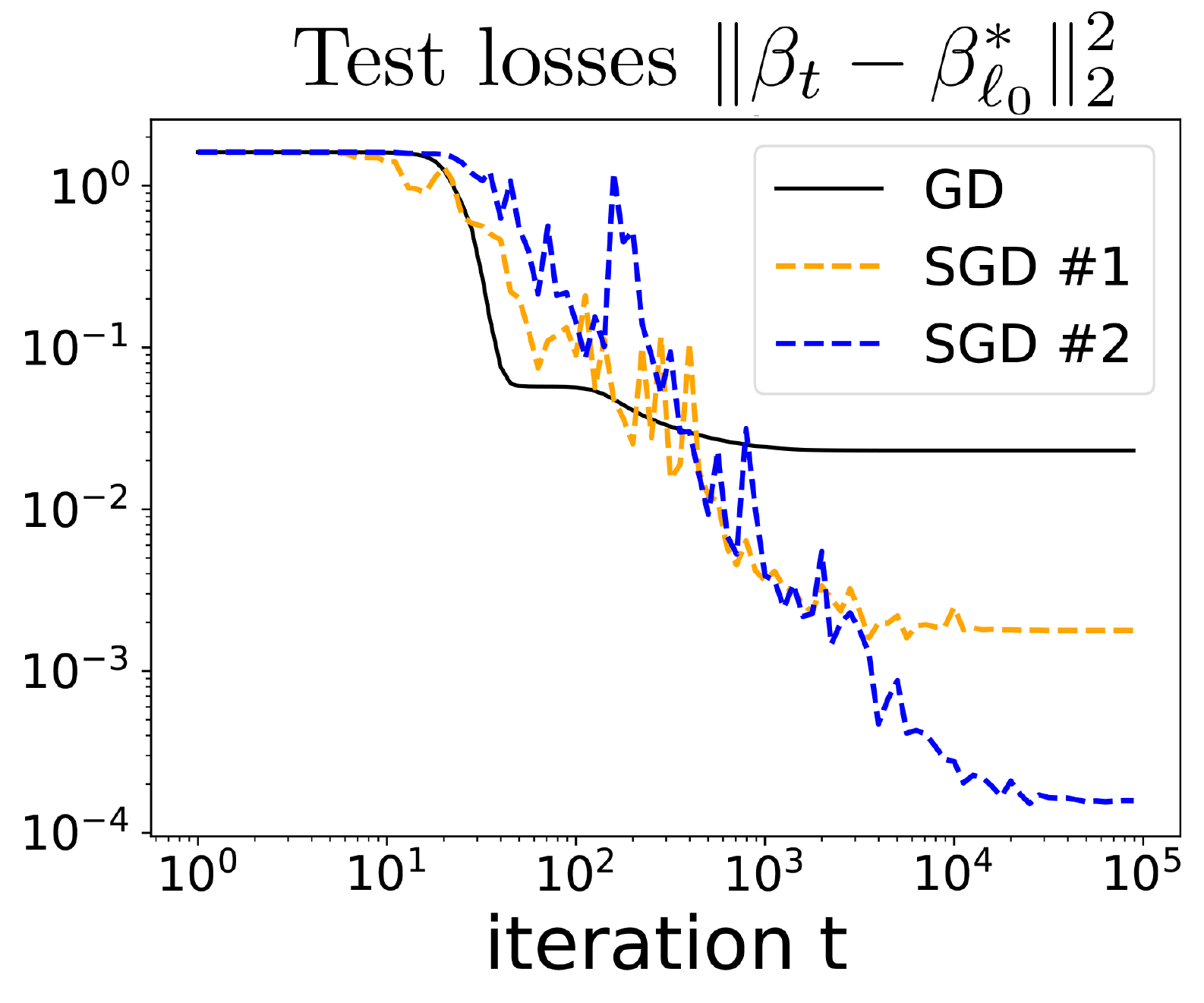}
   \end{minipage}
   \caption{ Sparse regression (see \Cref{subsection:experim_setup} for the detailed experimental setting). 
       Both SGD and GD are initialised at $\alpha = 0.1$.
   $2$ different runs of SGD over the training set are performed, they differ due to the inner stochasticity of the algorithm. 
    \textit{Left}: GD and SGD both converge towards a global minimum.
    \textit{Middle and right}: for two different trajectories of SGD, the higher the value of the loss integral at convergence, the better the validation loss. 
   In both cases SGD converges towards a solution which generalises better than GD.
 This figure illustrates \Cref{thm:main_theorem}.
     }
     \label{fig:main_theorem}
\end{figure}

%
%
\begin{restatable}{thm}{maintheorem}
\label{thm:main_theorem}
 For $ p \leq \frac{1}{2}$ and $w_{0, \pm} = \alpha  \in (\R_{+}^{*})^d$, let $(w_t)_{t \geq 0}$ follow the stochastic gradient flow~\eqref{eq:SGF} with step 
 size~$\gamma \leq O \big ( \big [\ln (\frac{4}{p})  \lambda_{\mathrm{max}} \max \{  \|\beta^*_{\ell_1}\|_{_1} \ln \big( \frac{ \Vert \beta^*_{\ell_1}  \Vert_1}{\min_i \alpha_i^2} \big), \|\alpha\|_2^2 \} \big ]^{-1}\big)$ 
 where \ \  $\beta^*_{\ell _1} = \underset{\beta \in \R^d\, \text{s.t.}\, X \beta = y}{ \argmin}\!\!\! \Vert \beta \Vert_1$
 and $\lambda_{\mathrm{max}}$ is the largest eigenvalue of  $X^{\top}X / n$.
Then, with probability at least~$1-p$:
\begin{itemize}
\item $(\beta_t)_{t \geq 0}$ converges towards a zero-training error solution $\beta_\infty^\alpha$
\item the solution $\beta^\alpha_\infty$ satisfies \vspace{0.5em}
\begin{equation}
\label{eq:main_theorem_min_prob}
       \hspace*{-1cm}\beta_\infty^\alpha = \underset{\beta \in \R^d \ \text{s.t.} \ X \beta = y}{ \mathrm{arg \ min \ }} \phi_{\alpha_{\eff}}(\beta) \quad \text{where } \ \ \alpha_{\eff} = \alpha \odot  \exp\left(\!- 2 \gamma \diag\left( \frac{X^{\top} X}{n} \right)\!\! \int_0^{+ \infty}\!\!\!\!\!\!\!\! L(\beta_s) \diff s \!\right).
\end{equation}
\end{itemize}
\end{restatable}
\setcounter{thm}{0}
The theorem is three-fold: with high probability and for an explicit choice of constant step size $\gamma$, 
(i) the flow $(\beta_t)_{t \geq 0}$ converges, 
(ii) its limit $\beta_\infty^\alpha$ is an interpolating solution, i.e. $X \beta_\infty^\alpha = y$ ,
(iii) this solution minimises the hyperbolic entropy problem with a parameter that depends on the dynamics.
We illustrate these results in \Cref{fig:main_theorem}.
Now let us comment further the theorem.

\vspace{-.5em}

\paragraph{Beneficial implicit bias through effective initialisation.} 
The most remarkable aspect of the result is that the recovered solution $\beta_\infty^\alpha$ minimises the same potential as for
 gradient flow but with an \emph{effective parameter} $\alpha_{\eff}$ which is strictly smaller than $\alpha$. 
 Hence, the hyperbolic entropy is closer to the $\ell_1$ norm compared to the deterministic case, proving a systematic 
 benefit of stochasticity. Note that this effective parameter is random and controlled by the loss integral $\int_0^{+ \infty} L(\beta_s) \diff s$: the higher the integral, the smaller the effective initialisation scale. In other words and quite surprisingly, the slower the loss converges to $0$, the ``richer'' the implicit bias.
However, it must be kept in mind that, as explained in \cite{woodworth2020kernel}, there is a tension between generalisation and optimisation: a 
longer training time might improve generalisation but comes at the cost of... a longer training time. 
Yet it is clear experimentally that
SGD systematically largely wins the trade-off over GD (see Figure~\ref{fig:main_theorem}).
Interestingly, Problem~(\ref{eq:main_theorem_min_prob}) tells us that the implicit bias of SGD initialised at $\alpha$ acts as if we run GD initialised at $\alpha_\eff$ 
(see \Cref{exp:GD_SGD}). 
Note that the minimisation problem~\eqref{eq:main_theorem_min_prob} only makes sense \textit{a posteriori} since the 
quantity $\alpha_\eff$ depends on the whole stochastic trajectory.
Finally, an interesting question is whether one can quantify the scale of this beneficial phenomenon, i.e. how small $\alpha_\eff$ is compared to $\alpha$. To answer this, we quantify the scale of the loss integral w.r.t. $\gamma$ and $\alpha$ (see Proposition~\ref{prop:loss_integral}) and show under slightly stronger conditions that the relative scale $\alpha_\eff/\alpha$ decays as power of $\alpha$ (See Eq.~\eqref{eq:scale_effective_initialisation} of the main text and Proposition~\ref{app:prop:true_scale_alpha_eff} of the appendix for a proof).

\paragraph{Kernel regime.} Though it is less our focus, our result still holds as $\alpha \to +\infty$ which corresponds to the kernel regime. 
In this regime, we believe that
$\int_0^{+ \infty} L(\beta_s) \diff s \underset{\alpha \to \infty}{\to} 0$ (not shown in the paper but experimentally 
observed) and hence SGF and GF converge towards the same solution.
This is expected since in the NTK regime, the iterates follow a kernel linear regression for which the bias of SGF and GF 
are the same. 

\paragraph*{Step size.}
Note that the convergence of the iterates holds for a constant step size. This is not illogical since in the overparametrised setting, the noise vanishes at the optimum (see \cite{varre2021iterate} for a convergence result in the overparametrised least-squares setup).
The explicit formula for the $\gamma$ upper bound is $\gamma \leq \Big (400 \ln \big( \frac{4}{p} \big ) \lambda_{\textrm{max}} (\frac{X^{\top} X }{n}) \max{ \Big \{  \Vert \beta^*_{\ell_1} \Vert_1 } 
\ln \big( \sqrt{2} \frac{ \Vert \beta^*_{\ell_1}  \Vert_1}{\min_i \alpha_i^2} \big)
, \Vert \alpha \Vert_2^2 \Big \}\Big)^{-1}$. It has a classical dependence
on $\lambda_{\mathrm{max}}(X^{\top} X/n )$ which can be computed, but also on the unknown value of $\Vert \beta^*_{\ell_1} \Vert_1$.
However in practice we choose the highest value of $\gamma$ for which the iterates converge. 
Note that in practice the weights are often  initialised 
such that $\Vert \alpha \Vert_2^2$ is roughly equal to $1$ and hence it is sensible to consider 
$\Vert \alpha \Vert_2^2 < \Vert \betasparse \Vert_1$.
In the explicit bound, there is a $\ln \big(  \Vert \betasparse \Vert_1/ \min_i \alpha_i^2 \big)^{-1}$ factor, we believe that it 
is an artefact of our analysis and could be removed.
It is hence best to think of the upperbound on $\gamma$ to simply be $\gamma \leq O( \frac{1}{\lambda_{\mathrm{max}} \Vert \betasparse \Vert_1 })$.

\paragraph{Convergence and proof sketch.} Let us put emphasis on the fact that since 
we deal with a non-convex problem, 
neither convergence nor convergence towards a global minimum are obvious. 
In most of similar works, convergence of the iterates is assumed \citep{woodworth2020kernel,gunasekar2018characterizing}.
In fact, the hardest and most technical part of our result is to show the convergence of the flow 
with high probability: once the convergence is shown, describing the minimisation problem $\beta_\infty^\alpha$ verifies 
is straightforward. In the following section we give several properties which constitute the major keys of the
theorem's proof.

\section{Links with mirror descent}
\label{sec:dynamical_properties}


The aim of this section is to show that the sequence $(\beta_t)_{t \geq 0}$ follows a stochastic version of continuous mirror descent with a time dependent mirror. From this crucial property, we show how the convergence and implicit bias characterisation follow. Finally, as it is one of the central objects of our main theorem, we give an estimation of $\int_0^\infty L(\beta_s) \diff s$.

\subsection{Stochastic continuous  mirror descent with time-varying potential}

We start by recalling known results on the link between implicit bias and mirror descent. We recall also convergence guarantees for mirror descent dynamics.

\paragraph{Mirror descent: convergence and implicit bias.} 

For any $\beta_0 \in \R^d$ and convex potential function $\Psi$, consider the mirror descent flow $(\beta_t)_t$ which corresponds to
$\diff \nabla \Psi(\beta_t) =  - \nabla L(\beta_t) \dd t$. 
Though the convergence of the loss to $0$ is straightforward, showing the convergence of the iterates 
requires more work and is shown in~\cite[Theorem~$2$]{bauschke2017descent} for strongly convex potentials.
 Yet, once the convergence of the iterates is shown, deriving the implicit minimisation problem is straightforward. 
 We recall the reasoning here~(see Section $3$ of \citep{azulay2021implicit} for more details): integrating the flow yields
$\nabla \Psi(\beta_\infty) - \nabla \Psi(\beta_0) = - \int_0^\infty \nabla L(\beta_s) \diff s =  
- 4 X^{\top} \int_0^\infty X (\beta_s - \beta_\infty ) \diff s \in \mathrm{span}(X)$. This condition, along 
with the fact that $X \beta_\infty = y$ exactly corresponds to the KKT conditions of the  problem:
\begin{align}
&\beta_{\infty} = \underset{\beta \in \R^d \ \text{s.t.} \ X \beta = y}{ \argmin } \ D_{\Psi}(\beta, \beta_0),
\end{align}
where $D_{\Psi}(\beta, \beta_0) = \Psi(\beta) - \Psi(\beta_0) - \langle \nabla \Psi(\beta_0), \beta - \beta_0 \rangle$ is the Bregman divergence w.r.t. $\Psi$.

\paragraph*{Link with our model.} It turns out that these general observations on mirror descent apply to our framework when $(w_t)_t$ follows the gradient flow 
$\diff w_{t, \pm} = - \nabla_{w_{\pm}} L(w_t) \diff t$. Indeed it has been shown in \cite{woodworth2020kernel}
that the corresponding iterates 
$\beta_t = w_{t,+}^2 - w_{t,-}^2$ follow a mirror descent with potential $\phi_\alpha$ defined in~Eq.\eqref{eq:potential_function}.
Therefore we can apply the previous remarks
 to obtain the convergence towards an interpolator\footnote{In our case, $\phi_\alpha$ is not strongly convex so a bit more work is necessary to show the convergence of the iterates (see \Cref{app:mirror_deterministic}).}, as well as the associated implicit 
minimisation problem which in our case can be rewritten as 
$\beta_{\infty}^\alpha = \underset{\beta \in \R^d \ \text{s.t.} \ X \beta = y}{ \argmin } \ \phi_\alpha(\beta)$ since 
$\nabla \phi_\alpha(\beta_0 = 0) = 0$.

\paragraph{Stochastic Mirror descent with a time varying potential.} 

To address the problem where $(w_t)_t$ follows a stochastic gradient flow 
instead of a gradient flow, it is natural, as in the deterministic framework,
to see what type of flow $(\beta_t)_t$ follows. 
Because of the noise, we cannot hope to simply recover a classical mirror descent. 
However interestingly the next property shows that it follows a 
stochastic mirror-like descent with a geometry that depends on time.

\begin{restatable}{prop}{stomirrordescent}
\label{prop:sto_mirror_descent}
Consider the iterates $(w_t)_{t \geq 0}$ issued from the stochastic gradient flow in~Eq.\eqref{eq:SGF} with initialisation $w_{0, \pm} = \alpha  \in (\R_{+}^{*})^d$.  
Then the corresponding flow $(\beta_t)_{t \geq 0}$ follows a ``stochastic continuous mirror descent with time varying potential'' defined by: 
\begin{align}
\label{sde:mirror_descent}
    \diff \nabla \phi_{\alpha_t}(\beta_t) = 
     - \nabla L(\beta_t) \diff t + \sqrt{\gamma n^{-1}L(\beta_t)} X^{\top} \mathrm{d}B_t,
\end{align}
where $\alpha_t = \alpha \odot \exp \big(\! -\! 2 \gamma \diag\left( \frac{X^{\top} X}{n} \right) \int_0^t L(\beta_s) \diff s \big)$ and $\phi_\alpha$ is the hyperbolic entropy defined in~\eqref{eq:potential_function}.
\end{restatable}

Under this form we clearly see that the iterates $(\beta_t)_t$ follow a flow which closely resembles that of mirror descent but 
with two major differences: 
(i) the potential $\phi_{\alpha_t}$ changes over time according to the random quantity $\int_0^t L(\beta_s) \diff s$, (ii) the flow is perturbed by noise. 
%
%
We highlight the fact that viewing the dynamics this way has the major advantage of giving 
a clear roadmap for the proof of Theorem~\ref{thm:main_theorem}:
(i) we can adapt classical mirror-descent results to our framework and construct appropriate Lyapunov functions to prove the convergence 
of the flow with high probability to some interpolator $\beta^\alpha_\infty$,
(ii) we immediately  recover the corresponding minimisation problem as in the deterministic case. 
Indeed, integrating Eq.\eqref{sde:mirror_descent} still yields
$\nabla \phi_{\alpha_{\eff}}(\beta_{\infty}^\alpha)  \in \mathrm{span}(X)$ which, along with $X \beta_{\infty}^\alpha = y$,
are the KKT conditions of the implicit minimisation problem \eqref{eq:main_theorem_min_prob}. 
We emphasise the fact that the structure of the noise, belonging to $\mathrm{span}(X)$, is crucial in order to obtain this minimisation problem. 
This would for instance clearly not be true 
if we considered isotropic noise in the SDE modelling.
This highlights the fact that not every form of noise improves the implicit bias: 
the shape of the intrinsic SGD noise is of primal importance~\cite{haochen2020shape}.



\subsection{Convergence and control of \texorpdfstring{$\int_0^\infty L(\beta_s) \diff s$}{PDFstring}}

Though it seems easy to derive the implicit minimisation problem \eqref{eq:main_theorem_min_prob} from the mirror-like structure of Eq.\eqref{sde:mirror_descent}, it is necessary to ensure that the iterates converge towards an interpolator $\beta_{\infty}$. 
This is the purpose of the following proposition.

\begin{prop}[Convergence of the iterates]
\label{prop:iterates_convergence}
Consider the iterates $(w_t)_{t \geq 0}$ issued from the stochastic gradient flow \eqref{eq:SGF}, initialised at $w_{0, \pm} = \alpha  \in (\R_{+}^{*})^d$.  
For $p \leq \frac{1}{2}$ and $\gamma$ such as in Theorem~\ref{thm:main_theorem}, then with probability at least $1 - p$, the flow $(\beta_t)_t$ converges to an interpolating solution $\beta_\infty^\alpha$.
\end{prop}
The convergence of the iterates is technical and requires several intermediate results. 
We start by considering an appropriate Bregman-type stochastic function with a time-varying potential and show that 
it converges with high probability.
Leveraging the fact that we are able to bound the iterates $\beta_t$, we are able to show 
that the limit of the function is in fact $0$. Owing to the fact that the function we 
consider also controls the distance of $\beta_t$ to a particular $\beta^*$ we finally get that 
the iterates converge.

However for the objects (such as $\alpha_\infty$) and functions we introduce to be well defined, we 
need to guarantee the convergence of  
$\int_0^\infty L(\beta_s)\dd s$. 
Besides, it is crucial to grasp the scale of this quantity 
since it gives the overall scale of $\alpha_\infty$.  This is done in the following proposition where we lower and upper bound 
its value.

%
%
\begin{restatable}{prop}{lossintegral}
\label{prop:loss_integral}
Under the same setting as in Proposition~\ref{prop:iterates_convergence} \
with initialisation $w_{0, \pm} = \alpha \mathbf{1}$, we have with 
probability at least $1 - p$: 
\begin{align*}
    \Omega \Big (    \Vert \betasparse \Vert_1   \ln \Big (  \frac{ \Vert \betasparse \Vert_1 }{\alpha^2}    \Big )   \Big )
      \underset{\alpha \to 0}{\leqslant}  
      \int_0^{+\infty} L(\beta_s) \diff s 
        \leqslant  O \Big (  \max \big \{  \Vert \betasparse \Vert_1   \ln  \Big (  \frac{ \Vert \betasparse \Vert_1 }{\alpha^2}   \Big ), \alpha^2 d \big \}   \Big ).
\end{align*} 
\end{restatable}
We point out that the lower bound is given for small $\alpha$'s for simplicity but we provide
in \Cref{app:lemma:B:gamma_int_loss} (\Cref{app:subsec:lowerbound_loss})
a lower bound which holds for all $\alpha$'s.
Note that when $\gamma = 0$, which corresponds to deterministic gradient flow, we can give the exact
value for the integral: $\int_0^{+\infty} L(\beta_s) \diff s = \frac{1}{2} D_{\phi_\alpha}(\beta_\infty^\alpha, \beta_0) $
(see \Cref{app:prop:mirror_descent} in \Cref{app:sec:deterministic}). 
This matches the scale of the bounds given in Proposition~\ref{prop:loss_integral}, hence showing the tightness of the result.
We focus now on how this translates to the scale of the effective initialisation w.r.t. $\alpha$ when this latter is small enough. In fact, this lower bound on the integral of the loss along with a stronger assumption 
on the boundedness of the iterates lead to 
\begin{align}
\label{eq:scale_effective_initialisation}
  \frac{\alpha_\infty}{\alpha}  \underset{\alpha \to 0}{\leqslant} \left (  \frac{\alpha^2}{\Vert \beta_{\ell_1}^* \Vert_1 } \right )^\zeta,
\end{align}
for some $\zeta > 0$. Hence  the smaller the initialisation scale $\alpha$ and the greater the 
benefit  of SGD over GD in terms of implicit bias (see \Cref{app:B:true_scale_alpha_inf} for more details).

Again, the proof of this  proposition is technical and relies on considering appropriate Lyapunov functions which 
highly resemble to Bregman divergences, but which take into account the fact that the geometry changes over time. These overall decreasing Lyapunov's enable to bound the iterates as well as lower and upper bound the integral of the loss. The stochastic integrals which naturally appear are controlled with high probability using time-uniform concentration of martingales~\citep{10.1214/18-PS321}.

\vspace{-.5em}
\section{Experiments}
\label{sec:going_further}

\subsection{Experimental setup for sparse regression}\label{subsection:experim_setup}

We consider the following sparse regression setup for our experiments. We choose $n = 40$, $d = 100$ and randomly 
generate a sparse model $\beta^*_{\ell_0}$ such that $\Vert \beta^*_{\ell_0}  \Vert_0 = 5$. We generate the features as $x_i \sim \mathcal{N}(0, I)$
and the labels as $y_i = x_i^{\top} \beta^*_{\ell_0}$. SGD, GD and the SGF are always initialised using the same scale $\alpha > 0$ and it is specified 
each time. We use the same step size for GD and SGD and choose it to be the biggest as possible why still ensuring convergence. Note
that since the true population covariance $\mathbb{E}[x x^{\top}] $ is equal to identity, the quantity  $\Vert \beta_t - \beta^*_{\ell_0} \Vert_2^2$ corresponds to the validation 
loss.

\subsection{Validation of the SDE model}

\begin{figure}[ht]
\centering
\hspace*{-20pt}
\begin{minipage}[c]{.45\linewidth}
\includegraphics[width=\linewidth]{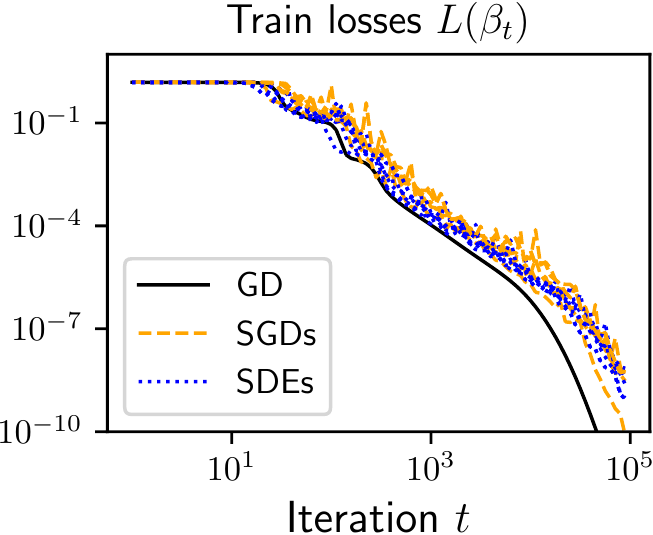}
   \end{minipage}
   \hspace*{10pt}
   \begin{minipage}[c]{.45\linewidth}
\includegraphics[width=\linewidth]{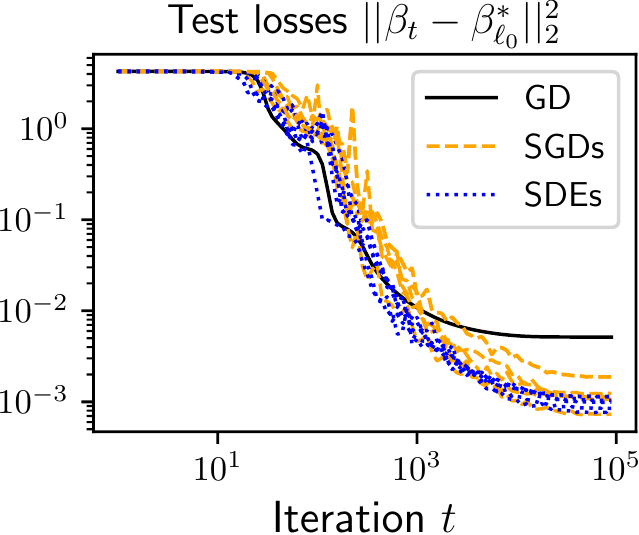}
   \end{minipage}
  \caption{Sparse regression (see \Cref{subsection:experim_setup} for the detailed experimental setup). 
  \textit{Left and right}: the training and the validation losses behave very similarly, corroborating the continuous modelling.}
     \label{fig:validation_experiments}
\end{figure}

In this section, we present an experimental validation of the stochastic gradient flow model. In Figure~\ref{fig:validation_experiments}, 
for the same step size, we run:
 (i)~the trajectory of gradient descent, (ii)~$5$ trajectories of stochastic gradient descent that correspond to different realisations of the uniform sampling over the data, 
 (iii)~$5$ trajectories of the stochastic gradient flow (its Euler discretisation with $\diff t = \gamma / 10)$) corresponding to different realisations of the Brownian. 
 We clearly see  (left) that the loss behaves similarly for SGD and SGF across time. We also see that the validation losses (right) of the iterates
 of SGD and SGF have very similar behaviours.
 This tends to validate our continuous modelling from \Cref{subsec:SDE_model}.

\vspace{-.5em}

\subsection{GD and SGD have the same implicit bias, but from different initialisations.}
\label{exp:GD_SGD}

\begin{figure}[h]
      \centering
      \hspace*{-20pt}
      \begin{minipage}[c]{.43\linewidth}
      \includegraphics[width=\linewidth]{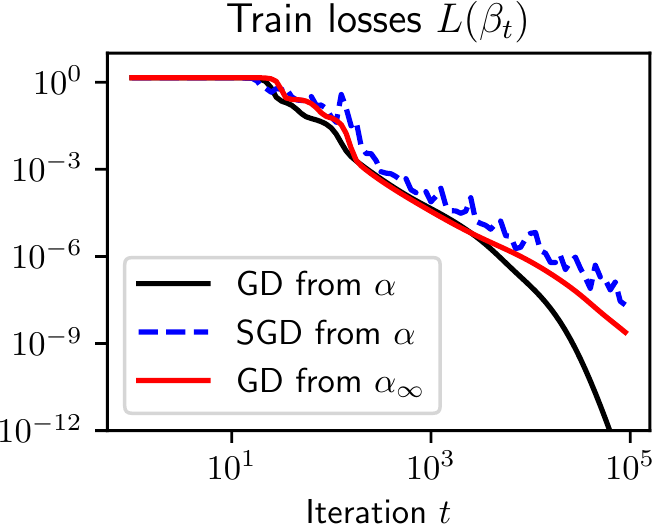}
         \end{minipage}
         \hspace*{10pt}
         \begin{minipage}[c]{.43\linewidth}
      \includegraphics[width=\linewidth]{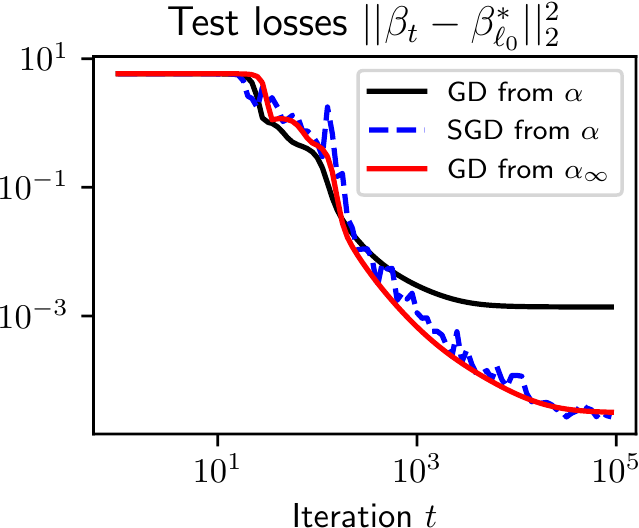}
         \end{minipage}
        \caption{Sparse regression (see \Cref{subsection:experim_setup} for the detailed experimental setup). 
        \textit{Left and right}: SGD initialised at $\alpha \mathbf{1}$
        converges towards the same point as
        GD initialised at $\alpha_\infty = \alpha \exp(- 2 \gamma \diag ( X^{\top} X / n) \int_0^\infty L(\beta_s^{\mathrm{SGD}}) \dd s)$ .}
           \label{fig:validation_theorem}
      \end{figure}

In order to confirm and illustrate the main \Cref{thm:main_theorem}, we provide the following experiment which is 
illustrated \Cref{fig:validation_theorem}.
We first run GD and SGD with the same step-size and initialise them both at $\alpha \mathbf{1}$ with $\alpha = 0.01$.
As expected, the solution recovered by SGD generalises better. Then, using the iterates $\beta_t^{\mathrm{SGD}}$ from the first SGD run, 
we compute the value $\alpha_\infty = \alpha \exp(- 2 \gamma \diag ( X^{\top} X / n) \int_0^\infty L(\beta_s^{\mathrm{SGD}}) \dd s) \in \R^d$ (the integral 
is approximated by its discrete time approximation with $\dd t = \gamma$). We then run gradient 
descent but this time initialised at $w_{0, \pm} = \alpha_\infty$. According to our main result 
from \Cref{thm:main_theorem}, it should approximately (it would be 
exact if we ran SGF and GF) converge to the same solution as SGD initialised at $\alpha \mathbf{1}$. This is clearly observed 
\Cref{fig:validation_theorem} (right). Also note that SGD and GD (initialised at $\alpha_\infty$) seem to have overall 
very similar dynamics, this is not shown by our results and we leave this as future work. However keep in mind that though the 
validation losses converge at the same iteration rate, in terms of computation time, SGD is $n$ times faster. 

\vspace{-.5em}

\subsection{Doping the implicit bias with label noise}
\label{main:subsec:label_noise}

As largely discussed throughout the paper, the effect of the implicit bias is controlled by the convergence speed of the loss: the slower it converges, the sparser the selected solution will be. 
Hence the following question: can we leverage this knowledge to dope the implicit bias? 
We argue in this Section that the answer to this question is affirmative. 
Indeed, consider a sequence $(\delta_t)_{t \in \mathbb{N}} \in \R_+^{\N}$ and assume 
that we artificially inject some label noise $\Delta_t$ at time $t$, say for example
$\Delta_t \sim \mathrm{Unif} \{2 \delta_t, -2 \delta_t \} $ (independently from $i_t$). 
This injected label noise perturbs the SGD recursion as follows:
\begin{align}
\label{eq:SGD_label_noise}
 w_{t+1, \pm} = w_{t, \pm} \mp \gamma  \left(\langle \beta_w - \beta^*, x_{i_t} \rangle + \Delta_t\right) \  x_{i_t} \odot w_{t, +}\ ,
 \qquad \  \text{where }\  i_t \sim \mathrm{Unif}(1, n).
\end{align}
As in Section~\ref{subsec:SDE_model}, we can derive its related stochastic gradient flow (see \Cref{app:subsec:label_noise} for more details):   
\begin{align}
 \label{eq:SGF_label_noise}
  \dd w_{t,\pm} &= - \nabla_{w_\pm} L(w_t) \dd t \pm 2 \sqrt{\gamma n^{-1} (L (w_t) + \delta_t^2)}\  w_{t,+} \odot [X^\top \dd  B_t].
 \end{align}
 Assuming that $(\delta_t)_{t \geq 0} \in (\R_+)^\R$ and $\gamma$ are such that the iterates  converge, 
 the corresponding implicit regularisation minimisation problem is preserved 
 but with a "slowed down" loss: $\tilde{L}(\beta_t):= L(\beta_t) + \delta_t^2$ and the effective initialisation writes: 
 $ \tilde{\alpha}_{\eff} = \alpha \odot \exp{\left(- 2 \gamma \diag(\frac{X^\top X}{n}) \int_0^{+ \infty} \tilde{L}(\beta_s)  \diff s\right)}$. 
The label noise therefore helps recovering a solution which has better 
sparsity properties.
However, it must be kept in mind that 
adding too much label noise can significantly slow down the convergence of the 
validation loss or even
prevent the iterates from converging.
Yet, experimental results showing the impressive effect of label noise are provided \Cref{app:fig:label_noise} in \Cref{app:subsec:label_noise}.



\section{Conclusion and Perspectives}

In this paper, we have shown the benefit of using stochastic gradient descent over gradient descent for diagonal linear networks in terms of their implicit bias.
 Indeed, we prove that stochastic gradient flow acts as gradient flow but initialised at a smaller scale: this induces a sparser finale iterate. This effect is controlled by the
 speed of convergence of the loss. Moreover, we prove the convergence of the flow and exhibit an interesting link with mirror descent. 
Fully understanding this novel type of dynamics could help to grasp the implicit biasing properties of stochastic gradient descent in other frameworks. 
It is also natural to ask whether the integral of the loss also controls the difference of implicit regularisation for more general architectures. 
It would also be interesting to analyse how this property adapts to $\log$ losses known to lead to max-margin solutions in classification. 

\paragraph{Acknowledgements.}
NF would like to thank Nathan Srebro for introducing him to the question of SGD's implicit bias as well as
for the stimulating discussions they had during his visit at EPFL.

\clearpage

\bibliographystyle{plainnat}
\bibliography{main_implicit_bias_SGD}

\clearpage



\clearpage

\appendix

%

{\bfseries \LARGE Appendix}

\paragraph*{Organisation of the Appendix.} 

The Appendix is structured as follows. 
In Section~\ref{app:SDE_modelling}, we give more precisions regarding the way we model stochastic gradient descent as a stochastic gradient flow. 
Section~\ref{app:proofs} is the core of the Appendix as it provides the proof of the theorem in a self-contained fashion.
For the sake of completeness, in Section~\ref{app:mirror_deterministic} we gather the results on the link between mirror-descent and implicit bias
 as well as give convergence results in the deterministic case (gradient flow).
In Section~\ref{app:experiments}, we provide more experiments supporting our results.
In Section~\ref{app:extensions}, we discuss some extensions of our results ; (\ref{app:subsection:E:general_SDE_model}) regarding a more general stochastic gradient flow model 
and in (\ref{app:subsec:depth_p}) we extend our results to depths $p\geq 3$.
Finally, Section~\ref{app:technical_lemmas} provides the technical material needed for the proofs of our results.


\section{Details on the SDE modelling}
\label{app:subsec:details_SDE_model}

We recall that the SGD recursion writes for $t \geqslant 1$ as:
\begin{equation*}
     \left.
        \begin{array}{ll}
            &w_{t+1, +} = w_{t, +} - \gamma  \langle \beta_w - \beta^*, x_{i_t} \rangle \  x_{i_t} \odot w_{t, +}  \\
            &w_{t+1, -} = w_{t, -} + \gamma  \langle \beta_w - \beta^*, x_{i_t} \rangle \  x_{i_t} \odot w_{t, -} 
        \end{array}
     \right. \qquad \text{where  } i_t \sim \mathrm{Unif}(1, n).
    \end{equation*}
Since the full gradient is $\nabla_{w_{\pm}} L(w) = \pm \big [\frac{1}{n} \sum_{k=1}^n \langle  \beta_w - \beta^*, x_{k} \rangle \  x_{k} \big ] \odot w_\pm \in \R^d$. We can 
rewrite the recursion as:
%
\begin{align*}
  w_{t+1, \pm} &= w_{t,\pm} - \gamma  \nabla_{w_{\pm}} L(w_t) \mp \gamma \Big [ \langle  \beta_{w_t} - \beta^*, x_{i_t} \rangle \  x_{i_t} - 
  \frac{1}{n} \sum_{k=1}^n \langle  \beta_{w_t} - \beta^*, x_{k} \rangle \  x_{k} \Big ] \odot w_{t, \pm}. 
\end{align*}
%
%
Now notice that 
\begin{align*}
  \langle  \beta - \beta^*, x_{i_t} \rangle \  x_{i_t} - 
  \frac{1}{n} \sum_{k=1}^n \langle  \beta - \beta^*, x_{k} \rangle \  x_{k}
&= X^{\top}
  \Big ( \langle \beta - \beta^*, x_{i_t} \rangle \mathbf{e}_{i_t} - \mathbb{E}_{i_t} \big [ \langle \beta - \beta^*, x_{i_t} \rangle \mathbf{e}_{i_t}  \big ] \Big ),
\end{align*}
where $\mathbf{e}_{i}$ is the $i^{th}$ element of the $\R^n$-canonical basis.
Let us denote by
$ \xi_{i_t}(\beta) = - \big ( \langle \beta - \beta^*, x_{i_t} \rangle \mathbf{e}_{i_t} - \mathbb{E}_{i_t} \big [ \langle \beta - \beta^*, x_{i_t} \rangle \mathbf{e}_{i_t}  \big ] \big )$. 
It is a zero-mean random variable with values in $\R^n$  and it can be seen as a multiplicative noise, i.e., proportional to $\beta-\beta^*$, which vanishes at the optimum.
The SGD recursion then writes as:
\begin{align*}
  w_{t+1, \pm} &= w_{t, \pm} - \gamma  \nabla_{w_{\pm}} L(w_t) \pm \gamma \big [ X^{\top} \xi_{i_t}(\beta_t) \big ] \odot w_{t, \pm} \\ 
     &= w_{t, \pm} - \gamma  \nabla_{w_{\pm}} L(w_t) \pm \gamma \diag(w_{t, \pm}) X^{\top} \xi_{i_t}(\beta_t).
\end{align*}
%
 As we are interested in the stochastic differential model of the SGD recursion, let  us now compute the covariance of the SGD noise. We first notice that 
\begin{align*}
  \text{Cov}_{i_t}[ \xi_{i_t}(\beta) ] &= \mathbb{E}_{i_t}[ \xi_{i_t}(\beta)^{\otimes 2} ] \\
  &= \mathbb{E}_{i_t}[ (\langle \beta - \beta^*, x_{i_t} \rangle \mathbf{e}_{i_t})^{\otimes 2} ] - \mathbb{E}_{i_t}[ \langle \beta - \beta^*, x_{i_t} \rangle \mathbf{e}_{i_t} ]^{\otimes 2} \\
  &= \frac{1}{n}
  \begin{pmatrix}
    \langle \beta - \beta^*, x_1 \rangle^2  &   &  0\\
      & \ddots & 0 \\
    0 &  & \langle \beta - \beta^*, x_n \rangle^2 \\
    \end{pmatrix}
    - \frac{1}{n^2}
    \Big ( \langle \beta - \beta^*, x_i \rangle \langle \beta - \beta^*, x_j \rangle \Big )_{1 \leq i, j \leq n} \\
      &= \frac{4}{n}
    \begin{pmatrix}
      L_1(\beta)  &   &  0\\
        & \ddots & 0 \\
      0 &  & L_n(\beta) \\
      \end{pmatrix} 
      - \frac{1}{n^2}
    \Big ( \langle \beta - \beta^*, x_i \rangle \langle \beta - \beta^*, x_j \rangle \Big )_{1 \leq i, j \leq n} 
\end{align*}
where 
$L_i(\beta) = \frac{1}{4} \langle \beta - \beta^*, x_i \rangle^2  $ is the individual loss of the observation  $x_i$, such that $L(\beta) = \frac{1}{n} \sum_{i = 1}^{n} L_i(\beta)$.

Thus, the covariance satisfies the relation 
$  \text{Cov}_{i_t}[ \xi_{i_t}(\beta) ] = \frac{4}{n} \diag (L_i(\beta) )_{1 \leq i \leq n} + O(\frac{1}{n^2})$.
From this expression we can obtain a good model for 
$\text{Cov}_{i_t}[ \xi_{i_t}(\beta) ] $. First, we neglect the second term of order $1/n^2$. 
Then, we assume that all partial losses are approximately uniformly equal to their mean: i.e. for any $i$, $\ L_i(\beta)\cong \E_{i_t} [L_{i_t}(\beta)] $ 
(the general case is discussed \Cref{app:subsection:E:general_SDE_model}). Hence,
\begin{align*}
     \text{Cov}_{i_t}[ \xi_{i_t}(\beta) ] 
     &\cong \frac{4}{n} \diag \Big ( \frac{1}{n} \sum_i L_i(\beta) \Big ) = \frac{4}{n} L(\beta) I_n.
\end{align*}
The overall SGD's noise structure is then captured by 
\begin{align*}
     \Sigma_{_{\mathrm{SGD}}} (w_\pm) 
     &:= \gamma^2 \diag(w_\pm) X^{\top} \text{Cov}_{i_t}[ \xi_{i_t}(\beta) ] X \diag(w_\pm) \\
&\cong \frac{4}{n} \gamma^2 L(\beta) [\diag(w_\pm) X^{\top} ]^{\otimes 2}.
\end{align*}

This leads us in considering the following SDE:
\begin{equation*}
    \begin{aligned}
    \dd w_{t,+} &= - \nabla_{w_+} L(w_t) \diff t + 2 \sqrt{\gamma n^{-1} L (w_t)}\  w_{t,+} \odot [X^\top \dd  B_t]  \\
    \dd w_{t,-} &= - \nabla_{w_-} L(w_t) \diff t - 2 \sqrt{\gamma n^{-1} L (w_t)}\  w_{t,-} \odot [X^\top \dd  B_t],
    \end{aligned}
    \end{equation*}
since its Euler discretisation with step size $\gamma$ is :
\begin{align*}
     w_{t+1, \pm} &= w_{t,\pm}  - \gamma \nabla_{w_\pm} L(w_t)  \pm 2 \sqrt{\gamma n^{-1} L (w_t)}\  w_{t,\pm} \odot [X^\top \varepsilon_t],
\end{align*}
where $\varepsilon_t \sim \mathcal{N}(0, \sqrt{\gamma} I_n)$. 
This corresponds to a Markov-Chain whose noise covariance is equal to $\Sigma_{_{\mathrm{SGD}}}$. 

\paragraph{Remark on mini-batch SGD.}
This analysis can easily be extended to a batch size larger than $1$.
Indeed, using a mini-batch sampled with replacement of size $b$
only changes the noise covariance up to a multiplicative constant as:
$\text{Cov}_{i_t}[ \xi^b_{i_t}(\beta) ] =\frac{1}{b} \text{Cov}_{i_t}[ \xi^{b' = 1}_{i_t}(\beta) ] $.
The associated SDE, for a step size $\gamma$, is therefore
$\dd w_{t,\pm} = - \nabla_{w_\pm} L(w_t) \diff t \pm 2 \sqrt{\gamma b^{-1} n^{-1} L (w_t)}\  w_{t,\pm} \odot [X^\top \dd  B_t]  $.
Hence, it the same SDE as for a batch-size equal to $1$ but with an effective step-size 
$\gamma_{\mathrm{eff}} = \gamma / b$ (hence larger step-sizes can be used, as expected). 
The exact same reasoning can be done for mini-batch without replacement and our analysis would hold this time with:
$\gamma_{\mathrm{eff}} = \gamma (n-b) / ((n - 1) b)$ . Note that all the results in our paper therefore
hold for mini-batch SGD by considering the effective step-size $\gamma_\mathrm{eff}$ instead of $\gamma$.


\label{app:SDE_modelling}


\section{Proofs of the main results}
\label{app:section:B}


This section contains all the proofs of the main results. It is self contained as we recall each time the propositions we prove. In subsection~\ref{app:subsec:Proposition_mirror}, 
we derive the mirror-descent-like flow which the iterates follow as in \Cref{prop:sto_mirror_descent} 
of the main text. Then, we upper bound the loss integral in subsection~\ref{app:subsec:upperbound_loss}. 
This leads us in proving the convergence of the iterates towards an interpolator in subsection~\ref{app:subsec:convergence_iterates}. 
Equipped with these results we prove the main result of the paper (\Cref{thm:main_theorem}) in subsection~\ref{app:subsec:main_theorem}. 
Finally, to complete the proof of~\Cref{prop:loss_integral} of the main text we derive a lower bound of the loss  in subsection~\ref{app:subsec:lowerbound_loss}.

For the sake of easy reading, we adopt the following notations in this section: we denote by $\tX := X / \sqrt{n}  $, and $\lambda_{\max} := \lambda_{\max}(\HH)$. 

\subsection{Proof of \texorpdfstring{\Cref{prop:sto_mirror_descent}}{proporthm}}
\label{app:subsec:Proposition_mirror}

In order to prove \texorpdfstring{\Cref{prop:sto_mirror_descent}}{proporthm}, we introduce the following lemma:

\begin{lemma}
    \label{app_lemma:closed_form_beta}
    Consider the iterates $(w_t)_{t \geq 0}$ issued from the stochastic gradient flow in~Eq.\eqref{eq:SGF} with initialisation $w_{0, \pm} = \alpha  \in (\R_{+}^{*})^d$.  
    Then we have the following implicit closed form expression for $\beta_t$:
    \begin{align}
    \label{eq:beta_implicit}
        \beta_t &= 2 \alpha_t^2  \odot \sinh ( 2 \tX^{\top} \eta_t ) ,
    \end{align}
where 
$\eta_t =  - \int_0^t  \tX (\beta_s - \beta^*) \diff s + 2 \sqrt{\tgamma } \int_0^t \sqrt{L(\beta_s)}  \dd B_s \in \R^n$
and
$\alpha_t = \alpha \odot  \exp \big  ( - 2 \tgamma \diag(\tX^{\top} \tX) \int_0^t L(\beta_s) \dd s \big )$.
 \end{lemma}
 Note that this \textbf{is not} an explicit closed form for $\beta_t$ since the right hand side  depends on $(\beta_s)_{0 \leq s \leq t}$.
\begin{proof}
Recall that the SDE we consider writes as: 
\begin{align*}
    \dd w_{t,\pm} &= - \nabla_{w_\pm} L(w_t) \diff t \pm 2 \sqrt{\gamma n^{-1} L (w_t)}\  w_{t,\pm} \odot [X^\top \dd  B_t]  \\
    &= \pm \big ( - [ \tX^{\top} r(w_t)] \odot w_{t, \pm} \diff t + 2 \sqrt{\tgamma  L (w_t)}\  w_{t,\pm} \odot [\tX^\top \dd  B_t] \big ),
\end{align*}
where 
$r(w) = \tX (w_+^2 - w_-^2 - \beta^*) = \tX (\beta_w - \beta^*)  \in \mathbb{R}^{n}$  are the (normalised) rests.

It turns out that there is an implicit closed form solution to this SDE. Indeed deriving the Itô formula on $\ln(w_{t,\pm})$ gives the following integral expression:
\begin{align*}
    w_{t, \pm}  & = w_{t = 0, \pm} \odot \exp ( \pm \tX^{\top} \Big [ - \int_0^t r(w_s) \diff s + 2 \sqrt{ \tgamma } \int_0^t \sqrt{L(w_s)}  \diff B_s \Big ]  )    \odot \exp ( - 2 \tgamma \diag(\tX^{\top } \tX)  \int_0^t L(w_s) \diff s ) \\
    &= \alpha_t \odot \exp ( \pm \tX^{\top} \eta_t  )   .
\end{align*}
Since $\beta = w_+^2 - w_-^2$, we get:
\begin{align*}
    \beta_t &= \alpha_t^2 \odot \big ( \exp ( + 2 \tX^{\top} \eta_t  ) - \exp ( - 2 \tX^{\top} \eta_t  ) \big ) \\
            &= 2 \alpha_t^2 \odot \sinh ( + 2 \tX^{\top} \eta_t  ). 
\end{align*}
\end{proof}

For clarity we recall the statement of \Cref{prop:sto_mirror_descent}.
\stomirrordescent*
\begin{proof}
The results immediately follows from \Cref{app_lemma:closed_form_beta}. Indeed, inverting the  implicit equation on~$\beta_t$, Eq.~\eqref{eq:beta_implicit}, we have,
\begin{align*}
     \arcsinh \Big(\frac{\beta_t}{2 \alpha_t^2}  \Big) 
     &= 2 X^{\top} \eta_t =  - 2 \tX^{\top} \int_0^t  \tX (\beta_s - \beta^*) \diff s + 4  \sqrt{\tgamma } \tX^{\top} \int_0^t \sqrt{L(\beta_s)}  \dd B_s.
\end{align*}
Hence,
\begin{align*}
    \diff \ \arcsinh \Big(\frac{\beta_t}{2 \alpha_t^2}\Big) 
    &=   - 2 \tX^{\top}   \tX (\beta_s - \beta^*) \diff t + 4  \sqrt{\tgamma } \tX^{\top}  \sqrt{L(\beta_t)}  \dd B_t \\
    &=   - 4 \nabla L(\beta_t)  \diff t + 4 \sqrt{\tgamma L(\beta_t)} \tX^{\top} \dd B_t.
\end{align*}

Noticing that $\nabla \phi_\alpha(\beta) = \frac{1}{4} \arcsinh (\frac{\beta}{2 \alpha^2}  ) $ concludes the proof.
\end{proof}

\subsection{Upperbound of the integral of the loss}
\label{app:subsec:upperbound_loss}

This section contains several technical arguments that permit us to derive the upperbound of the integral of the loss [\Cref{prop:loss_integral}, right side]. Let us try to highlight  the key features of this proof. 
First, as for classical mirror descent, we define a Lyapunov function that resembles a Bregman divergence plus  a necessary control term [Eq.~\eqref{app:eq_first_Lyapunov}]. 
Then, we fix a high-probability event on which we have a control of the Brownian diffusion term [Eq.~\eqref{app:eq:control_martingale}].
 This gives an equation involving a weighted integral of the loss. After lower bounding this weight to access directly the loss integral~[\Cref{app:lem:control_weight}], we show that the iterates themselves  are in fact bounded~[\Cref{lem:boundedness}]. 
 We finally conclude  the proof in \Cref{app:prop:loss_integral}.

\paragraph*{Notations and standard calculations.} Let us introduce some notations that are important throughout the proofs. We consider the hyperbolic entropy $\phi_\alpha(\beta)$ as a function of two variables $(y,z) \mapsto \phi(y, z)$ evaluated at the point $(\beta, \alpha^2) \in \R^d \times \R^d$. With a slight abuse of notation, we denote by $\nabla_\beta \phi(\beta,\alpha^2) \in \R^d$, the gradient with respect to the first vector evaluated in $(\beta, \alpha^2)$, and $\nabla_z \phi(\beta,\alpha^2) \in \R^d$, the gradient with respect to the second variable evaluated in $(\beta, \alpha^2)$. Let us also define the process $(\xi_t)_{t \geqslant 0}$,  as the vector  $\xi_t := \sqrt{\beta_t^2  + 4 \alpha_t^4} \in \R^d$, for all $t\geqslant0$. For the sake of clarity, we recall here the expression of the hyperbolic entropy as well as its derivatives: we have $\phi(\beta, \alpha^2) = \frac{1}{4}\sum_{i=1}^d \beta_i \ \arcsinh( \frac{\beta_i}{2 \alpha_i^2} ) - \sqrt{\beta_i^2 + 4 \alpha_i^4}$, and
\begin{align*}
&\nabla_{\beta} \phi(\beta, \alpha^2) =\frac{1}{4} \arcsinh\left(\frac{\beta}{2 \alpha^2}\right), \ \ \ \nabla_{z} \phi(\beta, \alpha^2) = - \frac{1}{4 \alpha^2} \sqrt{\beta^2 + 4 \alpha^4} \in \R^{d} \hspace*{0.5cm} \textrm{as well as,} \\ 
&\nabla^2_{\beta, \beta} \phi(\beta, \alpha^2) = \frac{1}{4} \diag \left[\frac{1}{\sqrt{\beta_i^2 + 4 \alpha_i^2}}\right]_i \in \R^{d \times d}.
\end{align*}

\paragraph*{A first Lyapunov function.} In this subsection we shall consider the following (stochastic) Lyapunov function:
\begin{align}
\label{app:eq_first_Lyapunov}
    V_t &:=  - 
    \phi_{\alpha_t}(\beta_{t})  +    \langle \nabla \phi_{\alpha_t}(\beta_t) , \beta_t - \betasparse \rangle +  \gamma \int^t_0 L(\beta_s) \dd s\  \langle |\betasparse|, \diag (\HH) \rangle.
\end{align}
This Lyapunov resembles to a Bregman divergence with respect to the hyperbolic entropy. The added term is however required to have a proper control on its decrease. Just as in the deterministic framework, we want to show that the Lyapunov is decreasing, i.e. 
it has a negative derivative. With this aim, we compute its Îto derivative $\dd V_t$ in the following lemma. 
\begin{lemma}
\label{app:lem:ito_V}
For all $t>0$, $V_t$ verifies the following equation:
    \label{app_lemma:Vt}
    \begin{align*}
        V_t &= V_0 - 2 \int_0^t L(\beta_s) \Big ( 1 - \frac{1}{2} \gamma   \langle \diag(\HH) , \xi_s + \vert \betasparse \vert \rangle \Big ) \dd s + \int_0^t \sqrt{ \tgamma L(\beta_s)} \langle X^{\top}   \dd B_s, \beta_s - \betasparse \rangle .
    \end{align*}
\end{lemma}
\begin{proof}
To derive the formula for the Lyapunov $V_t$, we compute its derivatives $\dd V_t$ thanks to Itô formula and then integrate it with respect to the time. Let us stress that as $V_t$ is a function of $\beta_t$ and $\alpha_t$ we need both their full Itô decomposition. For $\alpha_t$, as we know that $\alpha_t = \alpha \odot \exp \big  ( - 2 \gamma  \diag(\HH)  \int_0^{t} L(w_s)  \dd s \big ) $, we have $\dd \alpha_t = - 2 \gamma   \diag(\HH)   L(w_t) \alpha_t \dd t$. For $\beta_t$, we only need the noise compound of the Itô decomposition. Let us denote by $b(\beta_{w_t})$ the drift in the Itô decomposition of $\beta_t$\footnote{It can be computed but its precise formula is not needed.}, we have, 
\begin{align*}
\dd\beta_t &= \dd w^2_{t,+} - \dd w^2_{t,-} \\ 
 &= b(\beta_{w_t}) \dd t + 4 \sqrt{\gamma L (\beta_t)} (w_{t,+} \odot w_{t,+} \odot \left[\tX^{\top} \dd B_t\right] + w_{t,-} \odot w_{t,-} \odot \left[\tX^\top\dd B_t\right])  \\
 &= b(\beta_{w_t}) \dd t + 4 \sqrt{\gamma L (\beta_t)}\  \xi_t \odot \left[\tX^\top\dd B_t\right].
\end{align*}
From this expression, we deduce the matrix of its quadratic variations $\dd \langle \beta_t \rangle_{_\mathrm{qv}} = \left[\dd \langle \beta^i_t,  \beta^j_t\rangle \right]_{ij} = 16 \gamma L(\beta_t)(\tX^\top \tX) \odot (\xi_t \xi_t^\top)  \in \R^{d \times d}$.

We are now equipped to apply the Itô formula on $V_t$. Indeed, it is clear that $\phi$ is a $C^2$ function of $(\beta, \alpha)$, hence,
\begin{align*}
    \dd V_t 
    &= - \left[\langle \nabla_{\beta} \phi(\beta_{t}, \alpha_t^2) , \dd \beta_t \rangle  + \langle \nabla_{z} \phi(\beta_{t}, \alpha_t^2) , \dd \left[\alpha_t^2\right] \rangle + \frac{1}{2} \tr\left[ \nabla^2_{\beta,\beta} \phi(\beta_t, \alpha_t^2) \dd \langle \beta_t \rangle \right]\right] \\
    & \hspace*{2cm}+ \dd \left[\langle \nabla_\beta \phi(\beta_t, \alpha_t^2), \beta_t -\betasparse\rangle\right] +  \gamma L(\beta_t)  \langle |\betasparse|, \diag (\HH) \rangle \dd t.
\end{align*}
The fifth term is explicit. Let us treat the first four terms separately:

\underbar{\textit{First term}}. This term cancels  with a compound of the fourth term.

\underbar{\textit{Second term}}. We apply simply the chain rule for this term as $\alpha_t$ does not have any quadratic variation:
\begin{align*}
\langle \nabla_{z} \phi(\beta_{t}, \alpha_t^2) , \dd \left[\alpha_t^2\right] \rangle =  \left\langle -\frac{\xi_t}{4 \alpha_t^2} , 2\alpha_t \odot \dd \alpha_t \right\rangle =  \gamma L(\beta_t)\left\langle \xi_t,  \diag(\HH) \right\rangle\dd t.
\end{align*} 

\underbar{\textit{Third term}}. We directly see that $$\frac{1}{2}\tr\left[ \nabla^2_{\beta,\beta} \phi(\beta_t, \alpha_t^2) \dd \langle \beta_t \rangle \right] = \frac{1}{2}\tr\left[ \frac{1}{4}\diag\left(\frac{1}{\xi_t}\right) \cdot  4 \gamma  L(\beta_t) \HH \odot(\xi_t\xi_t^\top) \right] \dd t  =  2 \gamma  L(\beta_t)\langle \xi_t, \diag(\tX^{\top}\tX)\rangle \dd t. $$

\underbar{\textit{Fourth term}}. We apply Itô formula once again to get: 
\begin{align*}
  \dd \left[\langle \nabla_\beta \phi(\beta_t, \alpha_t^2), \beta_t -\betasparse\rangle\right] &= \langle \dd\left[\nabla_\beta \phi(\beta_t, \alpha_t^2)\right], \beta_t -\betasparse\rangle + \langle \nabla_\beta \phi(\beta_t, \alpha_t^2), \dd\beta_t\rangle +  \tr\left[ \dd \langle \nabla_\beta \phi(\beta_t, \alpha_t^2), \beta_t \rangle_{_\mathrm{vq}} \right],
\end{align*}
and thanks to Eq.~\eqref{sde:mirror_descent}, we have an expression for the first and last term, giving
\begin{align*}
  \dd \left[\langle \nabla_\beta \phi(\beta_t, \alpha_t^2), \beta_t -\betasparse\rangle\right] &= -\langle \nabla L(\beta_t), \beta_t -\betasparse\rangle \dd t + 2 \sqrt{\gamma L (\beta_t)} \langle \tX^\top \dd B_t, \beta_t -\betasparse\rangle  + \langle \nabla_\beta \phi(\beta_t, \alpha_t^2), \dd\beta_t\rangle \\
  &\hspace*{6cm}+ 4 \gamma L(\beta_t) \langle \xi_t, \diag(\tX^{\top}\tX) \rangle \dd t.
\end{align*}

\underbar{\textit{Final expression}}. Let us gather the four expressions to get $\dd V_t$. We remark that the terms  $\langle \nabla_\beta \phi(\beta_t, \alpha_t^2), \dd\beta_t\rangle$ cancels (from first and fourth terms) and since $\langle \nabla_\beta L(\beta_t), \beta_t -\betasparse\rangle = 2 L(\beta_t)$,
\begin{align*}
    \dd V_t 
    &=  - \left[ \gamma L(\beta_t)\left\langle \xi_t,  \diag(\HH) \right\rangle\dd t + 2 \gamma  L(\beta_t)\langle \xi_t, \diag(\tX^{\top}\tX)\rangle \dd t\right] - 2 L(\beta_t) \\ 
    &+  \sqrt{\gamma L (\beta_t)} \langle \tX^\top \dd B_t, \beta_t -\betasparse\rangle  + 4 \gamma L(\beta_t) \langle \xi_t, \diag(\tX^{\top}\tX) \rangle \dd t +  \gamma L(\beta_t)  \langle |\betasparse|, \diag (\HH) \rangle \dd t.
\end{align*}
And finally, we have the expression:
\begin{align*} 
    \dd V_t &= -2L(\beta_t) +  \gamma L(\beta_t)\left\langle \xi_t,  \diag(\HH) \right\rangle\dd t +   \gamma L(\beta_t)  \langle |\betasparse|, \diag (\HH) \rangle \dd t\\
    &\hspace*{6.5cm} + \sqrt{\gamma L (\beta_t)} \langle \tX^\top \dd B_t, \beta_t -\betasparse\rangle .
\end{align*}
Integrating this equation between $0$ and $t$ concludes the proof.  
\end{proof}

\paragraph*{Control of the martingale term and definition of $\mathcal{A}$.} \Cref{app_lemma:Vt} shows that in order to control~$V_t$, we need to
control the local martingale $S_t = \sqrt{ \gamma } \int_0^t  \sqrt{L(\beta_s)} \langle \tX^{\top}   \dd B_s, \beta_s - \betasparse \rangle $. In fact, it is expected that the deviation of $S_t$ from its quadratic variation is very small: this is a concentration property of local martingales similar to the Bernstein inequality for discrete ones~\cite{boucheron2013concentration}. 
To do so, let us fix $p < 1/2$ and we define two parameters: $a:=  \max \{ \Vert \beta_{\ell_1}^{*} \Vert_1 \ln (\sqrt{2} \frac{ \Vert \beta_{\ell_1}^{*} \Vert_1 }{\min\alpha_i^2}),   \|\alpha\|_2^2 \}$ and $b:= \frac{1}{2} \ln(4/p) a^{-1}$. The reason behind the precise value of $a$ will appear clearly in the proof of \Cref{lem:boundedness,app:lem:control_weight}. These parameters being fixed, we can define the  event:
\begin{align}
\label{app:eq:control_martingale}
\mathcal{A} = \{ \forall t \geq 0, \vert S_t \vert \leq a + 2 b \gamma  \lambda_{\max} \int_0^t  L(\beta_s)   (\Vert \beta_s \Vert_1^2 +  \Vert \beta_{\ell_1}^{*} \Vert_1^2)  \dd s\}.
\end{align} 
From Lemma~\ref{app:lemma:martingale_result}, we know that $\mathbb{P}(\mathcal{A}) \geq  1 - 2 \exp(- 2 a b ) = 1 - \frac{p}{2}$. Note that $p$ is a free parameter that can be chosen as small as we want. 

{\bfseries From now on and until the end of the Section, we place ourselves \textit{on} the event $\mathcal{A}$, that is, all (in)equalities between random variables should be considered pointwise for any~$\omega\in\mathcal{A}$. To make it clear, we will recall from time to time laconically this fact by writing, ``on $\mathcal{A}$''.}

From \Cref{app_lemma:Vt}, we deduce the following inequalities,
\begin{align*}
    V_t - V_0
    &\leq - 2 \int_0^t L(\beta_s) (1 - \frac{1}{2} \gamma   \langle \diag(\HH) , \xi_s +  \vert \betasparse \vert \rangle ) \dd s  + 2 b \gamma  \lambda_{\max} \int_0^t  L(\beta_s)  ( \Vert \beta_s \Vert_1^2 + \Vert \betasparse \Vert_1^2 ) \dd s + a \nonumber \\ 
    &\leq - 2 \int_0^t L(\beta_s) (1 - \frac{1}{2} \gamma   \langle \diag(\HH) , \xi_s +  \vert \betasparse \vert \rangle - b \gamma \lambda_{\max}  (\Vert \beta_s \Vert_1^2  + \Vert \betasparse \Vert_1^2   ) \dd s  + a.  \nonumber
\end{align*}
Hence, we have the following control on $V_t$ with respect to a weighted loss integral: 
    \begin{align}
    \label{app:eq:B:Vt}
     V_t - V_0 &\leq - 2 \int_0^t L(\beta_s) U_s \dd s + a,
\end{align}
where $U_t :=  1 - \frac{\gamma}{2} \big [   \langle \diag(\HH) , \xi_t + C \vert \betasparse \vert \rangle + 2 b \lambda_{\max}  (\Vert \beta_t \Vert_1^2  + \Vert \betasparse \Vert_1^2  )  \big ]  \leq 1$. The following lemma show that as long as $U_t$ stays positive, the iterates stay bounded.

\begin{lemma}
\label{lem:boundedness}
Let us place ourselves on the event $\mathcal{A}$. Let $\tau > 0$. Assume $(U_t)_{0 \leq t \leq \tau} $ is positive. 
Then for all $t \leqslant \tau $ we have the 
following explicit upper bound on both  $\Vert \beta_t \Vert_1$ and $ \Vert \xi_t \Vert_1$,
\begin{align*}
    \Vert \beta_t \Vert_1 \leq \Vert \xi_t \Vert_1  &\leq 18 \max \{ \Vert \beta_{\ell_1}^* \Vert_1 \ln (\sqrt{2} \frac{ \Vert \beta_{\ell_1}^* \Vert_1 }{\min\alpha_i^2}),   \|\alpha\|_2^2 \}.
\end{align*}
%
%
\end{lemma}

\begin{proof}
Let $t \leq \tau$.
Remember that $\alpha(t) = \alpha \odot \exp \Big  ( - 2 \gamma  \big ( \int_0^{t} L(w_s)  \dd s \big ) \diag(\HH)  \Big ) \in \R^d $.
Since $V_t \leq V_0 - 2 \int_0^t L(\beta_s) U(s) \dd s + a$ and since by assumption $U(s) \geq 0$ for all $s \leq t$, we immediately 
get that $V_t \leq V_0 + a = - \phi_\alpha(0) + a = \frac{1}{2} \|\alpha\|_2^2 + a$. Notice furthermore that $- \phi_{\alpha_t}(\beta_{t})  +    \langle \nabla \phi_{\alpha_t}(\beta_t) , \beta_t - \betasparse \rangle  = \frac{1}{4} \|\xi_t\|_1 - \frac{1}{4} \langle \arcsinh \frac{\beta_t}{2 \alpha_t^2}, \betasparse \rangle $. Hence, we have:
\begin{align*}
\Vert \xi_t \Vert_1 
&=   - 4\phi_{\alpha_t}(\beta_{t})  +   4 \langle \nabla \phi_{\alpha_t}(\beta_t) , \beta_t - \betasparse \rangle + \langle \arcsinh \frac{\beta_t}{2 \alpha_t^2}, \betasparse \rangle \\
&=  4 V_t  -
4 \gamma \int_0^t L(\beta_s) \dd s  \ \langle \vert \betasparse  \vert, \diag(\HH) \rangle + \langle \arcsinh \frac{\beta_t}{2 \alpha_t^2}, \betasparse \rangle \\
&\leq 2 \|\alpha\|_2^2 + 4 a + \langle \arcsinh \frac{\beta_t}{2 \alpha_t^2}, \betasparse \rangle -
4 \gamma \int_0^t L(\beta_s) \dd s  \ \langle \vert \betasparse  \vert, \diag(\HH) \rangle.
\end{align*}
We now use the fact that $\arcsinh(x) \leq \ln(2 ( x + 1) )$ and that $\vert x \vert + \vert y \vert \leq \sqrt{2} \sqrt{x^2 + y^2}$ for all $x,y\geq0$ .
\begin{align*}
\Vert \xi_t \Vert_1 
&\leq 2 \|\alpha\|_2^2 + 4 a + \sum_i \vert \beta_i^* \vert  \ln \left( \frac{\vert \beta_i(t) \vert + 2 \alpha_i(t)^2}{\alpha_i(t)^2} \right)    -
4 \gamma \int_0^t L(\beta_s) \dd s  \ \langle \vert \betasparse  \vert, \diag(\HH) \rangle \\
&\leq 2 \|\alpha\|_2^2 + 4 a + \sum_i \vert \beta_i^* \vert \ln \left(\sqrt{2}  \frac{\sqrt{ \vert \beta_i(t) \vert^2 + 4 \alpha_i(t)^4 }}{\min\alpha_i^2} \right)  - \sum_i \vert \beta_i^* \vert  \ln \left(\exp \Big  ( - 4 \gamma \int_0^{t} L(\beta_s)  \dd s  \diag(\HH)  \Big )\right)  \\
& \hspace*{8.2cm} -  4 \gamma \int_0^t L(\beta_s) \dd s  \ \langle \vert \betasparse  \vert, \diag(\HH) \rangle.   
\end{align*}
Since the last two terms cancel and for all $i$, $\sqrt{ \vert \beta_i(t) \vert^2 + 4 \alpha_i(t)^4 } \leqslant \|\xi\|_1$, we have
\begin{align*}
\Vert \xi_t \Vert_1 
&\leq   2 \|\alpha\|_2^2 + 4 a + \Vert \betasparse \Vert_{1}    \ln \left( \sqrt{2} \frac{ \Vert \xi_t \Vert_1 }{\min\alpha_i^2} \right).
\end{align*}
%
To obtain the explicit upperbound we use \Cref{app:lemma:F:A_B_log_bound} with 
 $A = \frac{2  \sqrt{2}\|\alpha\|^2}{\min\alpha_i^2} + \frac{4 a \sqrt{2}}{\min\alpha_i^2}$ and $B = \frac{ \sqrt{2} \Vert \betasparse \Vert_1 }{\min\alpha_i^2}$ 
since the condition on $A,B$ are  satisfied as $\frac{A}{B} + \ln(B) \geq  \frac{2 \|\alpha\|_2^2}{\Vert \betasparse \Vert_1} +  \ln(\sqrt{2} \frac{\Vert \betasparse \Vert_1 }{\min\alpha_i^2})  \geq 1 + \ln ( \sqrt{8} d )\geq 2 $, as soon as $d \geq 3$. Hence,
\begin{align*}
\Vert \beta_t \Vert_1\leq\Vert \xi_t \Vert_1  
&\leq \frac{5}{2} \left( 2\|\alpha\|_2^2 + 4 a  +  \Vert \betasparse \Vert_1  \ln\left(\frac{\sqrt{2}\Vert \betasparse \Vert_1}{\min\alpha_i^2}\right)\right) \\
  &\leq 3 \Vert \betasparse \Vert_1 \ln \left(\sqrt{2} \frac{ \Vert \betasparse \Vert_1 }{\min\alpha_i^2}\right) + 5 \|\alpha\|_2^2 + 10 a \\
  &\leq 18 \max \{ \Vert \beta_{\ell_1}^* \Vert_1 \ln (\sqrt{2} \frac{ \Vert \beta_{\ell_1}^* \Vert_1 }{\min\alpha_i^2}),   \|\alpha\|_2^2 \},
\end{align*}
where in the last inequality we plug in the value of $a$. This concludes the proof of the lemma.
\end{proof}

Recall that we defined $U_t =  1 - \frac{\gamma}{2} \big [   \langle \diag(\HH) , \xi_t + \vert \betasparse \vert \rangle + 2 b \lambda_{\max}  (\Vert \beta_t \Vert_1^2  + \Vert \betasparse \Vert_1^2  )  \big ] $. We now show that in fact $(U_t)_t$ is always lower bounded by a strictly positive constant. Hence,  the result of Lemma~\ref{lem:boundedness} is valid at any time $t>0$.

\begin{lemma}
\label{app:lem:control_weight}
On $\mathcal{A}$, let us fix $\gamma \leq [400 \lambda_{\max}  \ln ( \frac{4}{p} )  \max\{ \Vert \beta_{\ell_1}^{*} \Vert_1 \ln( \frac{\sqrt{2} \Vert \beta_{\ell_1}^{*} \Vert_1 }{\min\alpha_i^2} )  , \|\alpha\|_2^2 \}]^{-1}$. 
Recall that $U_t =  1 - \frac{\gamma}{2} \big [   \langle \diag(\HH) , \xi_t + \vert \beta_{\ell_1}^* \vert \rangle + 2 b \lambda_{\max}  (\Vert \beta_t \Vert_1^2  + \Vert \beta_{\ell_1}^* \Vert_1^2  )  \big ]$,
then for all $t \geq 0$,
$$U_t \geq \frac{1}{2}.$$
\end{lemma}
\begin{proof}
Let us define the stopping time $\tau = \inf \{ t \geq 0 \ \text{such that} \ U(t) \leq \frac{1}{2}  \}$. 
Note that 
\begin{align*}
    U_0 &=  1 - \frac{\gamma}{2} \big [   \langle \diag (\HH) , 2 \alpha^2 + \vert \betasparse \vert \rangle + 2 b \lambda_{\max}  \Vert \betasparse \Vert_1^2  )  \big ] \\
    &\geq 1 - \frac{\gamma}{2} \lambda_{\max}  \big [ 2 \|\alpha\|_2^2 +  \Vert \betasparse \Vert_1 + 2 b  \Vert \betasparse \Vert_1^2)  \\
    &\geq 1 - 2 \gamma \lambda_{\max}  a \ln( \frac{4}{p} ) \\
    &> \frac{1}{2},
\end{align*}
where the last inequality comes from the upperbound on $\gamma$.
Since $U_t$ is continuous we have that $\tau > 0$. Assume that $\tau < + \infty$, by definition of the stopping time, for $t \leq \tau$:
$U(t) \geq 0$ and we can apply \Cref{lem:boundedness} at time $\tau$:
\begin{align*}
    \Vert \beta_\tau \Vert_1 \leq \Vert \xi_\tau \Vert_1 \leq 18 \max \{ \Vert \betasparse \Vert_1 \ln \Big(\sqrt{2} \frac{ \Vert \betasparse \Vert_1 }{\min\alpha_i^2}\Big),   \|\alpha\|_2^2 \}.
\end{align*}
Therefore:
\begin{align*}
U_\tau &= 1 - \frac{\gamma}{2} \big [   \langle \diag (\HH) , \xi_\tau + \vert \betasparse \vert \rangle + 2 b \lambda_{\max}  (\Vert \beta_\tau \Vert_1^2  + \Vert \betasparse \Vert_1^2  )  \big ] \\
&\geq 1 - \frac{\gamma}{2} \lambda_{\max} \big [  \Vert \xi_\tau \Vert_1  +  \Vert \betasparse \Vert_1 + 2 b  (\Vert \beta_\tau \Vert_1^2  + \Vert \betasparse \Vert_1^2  )  \big ]  \\
&\geq 1 - \frac{\gamma}{2} \lambda_{\max} \Big [ 18    \max\{ \Vert \betasparse \Vert_1 \ln \Big( \frac{\sqrt{2} \Vert \betasparse \Vert_1 }{\min\alpha_i^2} \Big)  , \|\alpha\|_2^2 \}    \\
& \hspace*{3cm} + 2 \cdot 18^2 \cdot b \max\{ \Vert \betasparse \Vert_1^2 \ln^2 \Big( \frac{\sqrt{2} \Vert \betasparse \Vert_1 }{\min\alpha_i^2} \Big) , \|\alpha\|_2^4 \}   \Big ]. 
\end{align*}
Since $b = \frac{1}{2} \ln( \frac{4}{p} ) \max\{ \Vert \betasparse \Vert_1 \ln( \frac{\sqrt{2} \Vert \betasparse \Vert_1 }{\min\alpha_i^2} )  , \|\alpha\|_2^2 \}^{-1}  $  we get that:
\begin{align*}
    U_\tau 
    &\geq 1 - \frac{\gamma}{2} \lambda_{\max} \ln( \frac{4}{p} )  \max\{ \Vert \betasparse \Vert_1 \ln\Big( \frac{\sqrt{2} \Vert \betasparse \Vert_1 }{\min\alpha_i^2} \Big)  , \|\alpha\|_2^2 \}  
     [ 18 + 18^2  ] \\
&\geq 1 - 175 \ln (\frac{4}{p}) \gamma  \lambda_{\max} \max\{ \Vert \betasparse \Vert_1 \ln\Big( \frac{\sqrt{2} \Vert \betasparse \Vert_1 }{\min\alpha_i^2} \Big)  , \|\alpha\|_2^2 \} \\
&> \frac{1}{2},
\end{align*}
where the last inequality comes from the choice of $\gamma$.
 
This is inconsistent since $U_{\tau} = \frac{1}{2}$. Hence $\tau = + \infty$ and thus $U_t \geq 1/2$ for all $t$.
\end{proof}
From the result of Lemma~\ref{app:lem:control_weight}, with \Cref{app:eq:B:Vt}, we obtain:
\begin{align}
    \label{app:B:eq:loss_upperbound}
    \int_0^t L(\beta_s) \dd s  &\leq V_0 - V_t + a \leq 
        - V_t + 2 \max\{ \Vert \betasparse \Vert_1 \ln\Big( \frac{\sqrt{2} \Vert \betasparse \Vert_1 }{\min\alpha_i^2} \Big)  , \|\alpha\|_2^2 \}.
\end{align}
Hence it remains to lower bound $V_t$ in order to get the convergence of the integral of the loss.
\begin{lemma} 
\label{app:B:lemma:lower_bound_Vt}
    On $\mathcal{A}$, let $\gamma$ be set as in \Cref{lem:boundedness}, for  all $t> 0$, we have the following lower bound on~$V_t$:
$$ V_t \geq - \frac{\Vert \beta_{\ell_1}^* \Vert_1}{4}  \ln \Big ( \frac{18 \sqrt{2} }{\min\alpha_i^2} \max\left\{  \Vert \beta_{\ell_1}^* \Vert_1 \ln \Big(\sqrt{2} \frac{ \Vert \beta_{\ell_1}^* \Vert_1 }{\min\alpha_i^2}\Big),    \|\alpha\|^2_2 \right\} \Big).$$
\end{lemma}

\begin{proof}
We follow exactly the same proof as for upperbounding the iterates.
\begin{align*}
4 V_t
&=  \sum_i \sqrt{ \beta_i^2 + 4 \alpha_i(t) ^4} -  \langle \arcsinh \frac{\beta_t}{2 \alpha_t^2}, \betasparse \rangle + 
4 \gamma  \int_0^t L(\beta_s) \dd s  \ \langle \vert \betasparse\vert ,  \diag H \rangle \\
&\geq \Vert \xi_t \Vert_1 - \sum_i \vert \beta_i^* \vert  \ln \left(  \frac{\vert \beta_i(t) \vert + 2 \alpha_i(t)^2}{\alpha_i(t)^2} \right)    +
4 \gamma \int_0^t L(\beta_s) \dd s  \ \langle \vert \betasparse\vert,   \diag H \rangle \\
&\geq \Vert \xi_t \Vert_1 - \Vert \betasparse \Vert_1 \ln \Big( \sqrt{2} \frac{\Vert \xi_t \Vert_1}{\min\alpha_i^2} \Big)  \\
&\geq - \Vert \betasparse \Vert_1 \ln \Big( \sqrt{2} \frac{\Vert \xi_t \Vert_1}{\min\alpha_i^2} \Big)  \\
&\geq - \Vert \betasparse \Vert_1  \ln \Big ( \frac{18 \sqrt{2} }{\min\alpha_i^2} \max\left\{  \Vert \betasparse \Vert_1 \ln \Big(\sqrt{2} \frac{ \Vert \betasparse \Vert_1 }{\min\alpha_i^2}\Big),    \|\alpha\|^2_2 \right\} \Big).
\end{align*} 
\end{proof}

Hence $(V_t)_{t \geq 0}$ is lowerbounded and we can derive an upper bound on the loss integral to show  the right part of \Cref{prop:loss_integral}. We recall it here in the  following  proposition.  

\begin{prop}
\label{app:prop:loss_integral}
On $\mathcal{A}$, let $\gamma$ be set as in \Cref{lem:boundedness}, we have the following upper bound on the loss integral:
$$\forall t >0, \hspace*{0.5cm} \int_0^t L(\beta_s) \dd s \leq \tilde{O}\left(\max\left\{ \Vert \beta_{\ell_1}^{*} \Vert_1 \ln\left( \frac{ \Vert \beta_{\ell_1}^{*} \Vert_1 }{\min\alpha_i^2} \right)  , \|\alpha\|_2^2 \right\}\right).$$
As a consequence, the  integral $\int_0^\infty L(\beta_s) \dd  s$ converges.
\end{prop}

\begin{proof}
From \Cref{app:B:eq:loss_upperbound}, we have that
\begin{align*}
    \int_0^t L(\beta_s) \dd s &\leq 
        - V_t + 2 \max\{ \Vert \betasparse \Vert_1 \ln\Big( \frac{\sqrt{2} \Vert \betasparse \Vert_1 }{\min\alpha_i^2} \Big)  , \|\alpha\|_2^2 \},
\end{align*}
and  thanks to the lower bound on $V_t$ from \Cref{app:B:lemma:lower_bound_Vt}, it yields,
\begin{align*}
    \int_0^t L(\beta_s) \dd s &\leq 
       \frac{\Vert \betasparse \Vert_1}{4}  \ln \Big ( \frac{18 \sqrt{2}}{\min\alpha_i^2} \max\left\{  \Vert \betasparse \Vert_1 \ln \Big(\sqrt{2} \frac{ \Vert \betasparse \Vert_1 }{\min\alpha_i^2}\Big),    \|\alpha\|^2_2 \right\} \Big) + 2 \max\{ \Vert \betasparse \Vert_1 \ln\Big( \frac{\sqrt{2} \Vert \betasparse \Vert_1 }{\min\alpha_i^2} \Big)  , \|\alpha\|_2^2 \},
\end{align*}
hence  the  integral $\int_0^\infty L(\beta_s) \dd  s$ converges and we have furthermore the $\tilde{O}$ bound of the proposition.
\end{proof}

\subsection{Proof of the convergence of the iterates: \texorpdfstring{\Cref{prop:iterates_convergence}}{PDFstring}}
\label{app:subsec:convergence_iterates}

In this subsection we prove the convergence of the iterates which corresponds to \Cref{prop:iterates_convergence} of the main text. For the sake of completeness, we recall this fact in the following lemma.

\begin{lemma}
    \label{app:lemma:B:convergence_iterates}
On $\mathcal{A}$, 
let $\gamma \leq [400 \lambda_{\max}  \ln ( \frac{4}{p} )  \max\{ \Vert \beta_{\ell_1}^{*} \Vert_1 \ln( \frac{\sqrt{2} \Vert \beta_{\ell_1}^{*} \Vert_1 }{\min\alpha_i^2} )  , \|\alpha\|_2^2 \}]^{-1}  $. The iterates $(\beta_t)_{t \geq 0}$ converge to an interpolator $\beta_{\infty}^\alpha$, i.e. such that  $L(\beta_{\infty}^\alpha) = 0$.
\end{lemma}
\begin{proof}
Consider the following Bregman divergence style function for any interpolator $\beta^*$ :
\begin{align*}
    W_t = \phi_{\alpha_{\infty}}(\beta^* ) - \phi_{\alpha_{t}}(\beta_{t}  )  + 
\langle \nabla \phi_{\alpha_t}(\beta_t) , \beta_t - \beta^* \rangle, 
\end{align*}
where 
$\alpha_\infty = \alpha  \exp \Big  ( - 2 \gamma  \big ( \int_0^\infty L(\beta_s)  \dd s \big ) \diagH  \Big ) > 0$ is well 
defined on $\mathcal{A}$ as a result of \Cref{app:prop:loss_integral}. 
The exact same computations as in \Cref{app:lem:ito_V} lead to:
\begin{align*}
    W_t  & = W_0 - 2 \int_0^t L(\beta_s) \dd s + \langle \diagH,  \gamma \int_0^t   L(\beta_s)  \xi_s \dd s  \rangle  +  \sqrt{ \gamma } \int_0^t \sqrt{L(\beta_s)} \langle X^{\top}   \dd B_s, \beta_s - \beta^* \rangle.
\end{align*}
Note that:
\begin{itemize}
    \item $ \int_0^t L(\beta_s) \dd s $ converges from \Cref{app:prop:loss_integral}.
    \item $\int_0^t  \Vert L(\beta_s)  \xi_s \Vert_1 \dd s \leq \max_{s \geq 0} ( \Vert \xi_s \Vert_1 ) \int_0^t  L(\beta_s)  \dd s < \infty $ from \Cref{app:prop:loss_integral}. Hence $\int_0^t  L(\beta_s)  \xi_s \dd s$ is absolutely convergent, hence converges.
    \item $\int_0^t \sqrt{L(\beta_s)} \langle X^{\top}   \dd B_s, \beta_s - \beta^* \rangle$ has a quadratic variation equal to $ 4 \int_0^t L(\beta_s)^2 \dd s $ 
    and $ 4 \int_0^t L(\beta_s)^2 \dd s \leq  2 \lambda_{\max} \int_0^t  L(\beta_s)   (\Vert \beta_s \Vert_2^2 +  \Vert \beta^* \Vert_1^2) \dd s $.  This implies that the quadratic variation converges.  Hence we obtain the convergence~\footnote{See for example Theorem $5$ of \url{https://almostsuremath.com/2010/04/01/continuous-local-martingales/} for a proof of this fact. For the moment we did not find a precise reference of this standard fact in the classical~\cite{revuz2013continuous}.} of the Brownian integral $\int_0^t \sqrt{L(\beta_s)} \langle X^{\top}   \dd B_s, \beta_s - \beta^* \rangle$.
\end{itemize}
Overall we get that $W_t$ converges for all choice of interpolator $\beta^*$. 
Now note that since $\int_0^\infty L(\beta_s) \dd s < + \infty$ we can extract a subsequence such that $L(\beta_{\phi(t)}) \underset{t \to \infty}{\to} 0$. 
Since $(\beta_t)_t$ is bounded (\Cref{lem:boundedness,app:lem:control_weight}), so 
is  $(\beta_{\phi(t)})_t$ and we can extract a new subsequence which converges. Let $\beta_{\infty}^\alpha$ denote the limit:
$\beta_{\phi_2(t)} \underset{t \to \infty}{\longrightarrow} \beta_\infty^\alpha$ where $\phi_2$ is the double extraction.
Since $L(\beta_{\phi(t)}) \underset{t \to \infty}{\to} 0$ so does 
$L(\beta_{\phi_2(t)}) \underset{t \to \infty}{\to} 0$. By continuity of the loss 
we have that $\beta_\infty^\alpha$ is an interpolator.
Now notice that since the Lyapunov $W_t$ with the choice $\beta^* = \beta_{\infty}$
converges and that 
$W_{\phi_2(t)} \underset{t \to \infty}{\to}  0$ we get that  $W_t \underset{t \to \infty}{\to}  0$. 

Furthermore:
\begin{align*}
    W_t &= \phi_{\alpha_\infty}(\beta_\infty^\alpha) - \phi_{\alpha_t}(\beta_{t} )  + 
\langle \nabla \phi_{\alpha_t}(\beta_t) , \beta_t - \beta_\infty^\alpha \rangle  \\
&\geq \phi_{\alpha_t}(\beta_\infty^\alpha) - \phi_{\alpha_t}(\beta_{t})  + 
\langle \nabla \phi_{\alpha_t}(\beta_t) , \beta_t - \beta_\infty^\alpha \rangle  \\
&= D_{\phi_{\alpha_t}}(\beta_\infty^\alpha, \beta_t  ) \\
&\geq 0
\end{align*}
where the first inequality is because $\alpha \mapsto \phi_\alpha(\beta) $ is decreasing and $\alpha_t \geq \alpha_\infty$.
Therefore $D_{\phi_{\alpha_t}}(\beta_\infty^\alpha, \beta_t  ) \to 0$. Finally, since:
\begin{align*}
\nabla^2 \phi_{\alpha_t}(\beta_t) &= \text{diag} ( \frac{1}{\sqrt{\beta_i(t)^2 + 4 \alpha_t^4(i)}})_i \\
&\geq \text{diag} ( \frac{1}{\sqrt{  \max_s \{ \beta_i(s)^2 \} + 4 \alpha^4}})_i \\
&\geq   \text{diag} ( \frac{1}{\sqrt{  \max_s \{ \Vert \beta(s) \Vert_1^2 \} + 4 \alpha^4}})_i \\
&\geq \mu I_d,
\end{align*}
for some $\mu$ since the iterates are bounded. Therefore  for all $t \geq0$, $\phi_{\alpha_t}$ is $\mu$-strongly convex on some convex set in which the iterates $\beta_s$ stay in.
Which means that: 
$D_{\phi_{\alpha_t}}(\beta_\infty^\alpha, \beta_t  ) \geq \frac{\mu}{2} \Vert \beta_t - \beta_\infty^\alpha \Vert_2^2$. Hence $\beta_t \to \beta_\infty^\alpha$.
\end{proof}

\Cref{app:lemma:B:convergence_iterates} along with the fact that the event $\mathcal{A}$ 
has probability at least $1-\frac{p}{2}$ (see \Cref{app:lemma:martingale_result} and paragraph around \ref{app:eq:control_martingale}) concludes the proof of \Cref{prop:iterates_convergence}.

\subsection{Proof of \texorpdfstring{\Cref{thm:main_theorem}}{f}}
\label{app:subsec:main_theorem}

We are now equipped to prove the main result of the paper. For clarity we recall the statement of the theorem here.
\maintheorem*

\begin{proof}
Recall first that on $\mathcal{A}$, \Cref{app:lemma:B:convergence_iterates} implies that the iterates converge 
towards a zero-training error we denote by $\beta_{\infty}^\alpha$.
From \Cref{prop:sto_mirror_descent} we also have that:
\begin{align}
\label{app:equation:MD:SGF}
        \diff \nabla \phi_{\alpha_t}(\beta_t) = 
         - \nabla L(\beta_t) \diff t + \sqrt{\tgamma L(\beta_t)} \tX^{\top} \mathrm{dB}_t,
\end{align}
where $\alpha_t = \alpha \odot \exp \big(\! -\! 2 \tgamma \diag\left( \tX^{\top} \tX \right) \int_0^t L(\beta_s) \diff s \big)$ and $\phi_\alpha$ is the hyperbolic entropy defined in~\eqref{eq:potential_function}.
Since the quantity $\int_0^\infty L(\beta_s) \diff s $ is well defined on $\mathcal{A}$ (\Cref{app:prop:loss_integral}), we can integrate  
\eqref{app:equation:MD:SGF} from $t = 0$ to $t = \infty$
 which leads to $\nabla \phi_{\alpha_{\eff}} (\beta_{\infty}^\alpha) \in \mathrm{span}(X)$.
This condition, along with the fact that $X \beta_{\infty}^\alpha = y$, exactly corresponds to 
the KKT conditions of the implicit minimisation problem \eqref{eq:main_theorem_min_prob}. 
From~\Cref{app:lemma:martingale_result}, the fact that the event $\mathcal{A}$ 
has probability at least $1-p$ concludes the proof.
\end{proof}

\subsection{Lower bound on \texorpdfstring{$\int L(\beta_s) \dd s$}{PDFstring} and proof of Proposition~\texorpdfstring{\ref{prop:loss_integral}}{fez}}
\label{app:subsec:lowerbound_loss}

Similarly to what has been done in subsection~\ref{app:subsec:upperbound_loss}, in order to lower bound the loss integral, 
we need a (different) control on the deviation of the local martingale $S_t$. 
We choose $\hat{a} := W_0^\alpha/2$ and $\hat{b}:= \frac{1}{2} \ln(4/p) \hat{a}^{-1}$ so that once again $\hat{a}\hat{b} = \frac{1}{2} \ln(4/p)$. We refer to \Cref{app:lemma:B:gamma_int_loss} for the definition of $W_0^\alpha$. Now that these parameters are fixed, consider the new event: 
$$\mathcal{B} = \{ \forall t \geq 0, \vert S_t \vert \leq \hat{a} + 2 \hat{b} \gamma  \lambda_{\max} \int_0^t  L(\beta_s)   (\Vert \beta_s \Vert_1^2 +  \Vert \beta_{\ell_1}^{*} \Vert_1^2)  \dd s\}$$ 
In this entire subsection we shall put ourselves on the intersection $\mathcal{A} \cap \mathcal{B}$ which occurs with probability $\P(\mathcal{A}\cap \mathcal{B}) \geq 1 - (\P(\mathcal{A}^{C}) + \P(\mathcal{B}^C)) \geq 1 - p$.
Furthermore since the goal of this section is to obtain an idea of the dependency on $\alpha$ of the integral 
of the loss as $\alpha$ goes to $0$, we shall consider the initialisations $\alpha = \alpha \mathbf{1}$, therefore 
for now on $\alpha$ is a positive scalar. Note that with this convention $\|\alpha\|^2_2 = \alpha^2 d$.

Notice that the quantity $\gamma \int_0^{+ \infty} L(\beta_s) \dd s$, through $\alpha_\infty$, controls the 
magnitude of the sparse-inducing effect. 
In the following lemma we show that this quantity is lower bounded by a quantity which is strictly increasing with 
$\gamma$. \textbf{This recommends to pick the largest $\gamma$
(as long as the iterates converge). This fact is also observed in practice.}

\begin{lemma}
    \label{app:lemma:B:gamma_int_loss}
On $\mathcal{A}\cap\mathcal{B}$, let    $\gamma \leq [400 \lambda_{\max}  \ln ( \frac{4}{p} )  \max\{ \Vert \beta_{\ell_1}^{*} \Vert_1 \ln( \frac{\sqrt{2} \Vert \beta_{\ell_1}^{*} \Vert_1 }{\alpha^2} )  , \alpha^2 d \}]^{-1}  , $
\begin{align*}
    \gamma \int_0^{+ \infty} L(\beta_s) \dd s 
    &\geq \frac{W_0^\alpha}{4} \frac{\gamma }{1 + \gamma \frac{M}{W_0^\alpha}  },
\end{align*}
where
$ W_0^{\alpha} = \underset{\beta \ \text{s.t} \ X \beta = Y}{\text{min}} \phi_\alpha(\beta) - \phi_\alpha(0) $ 
and  
\mbox{$M = \big [ 325  \lambda_{\max}  \ln ( \frac{4}{p} )  \max\{ \Vert \beta_{\ell_1}^{*} \Vert_1^2 \ln^2( \frac{\sqrt{2} \Vert \beta_{\ell_1}^{*} \Vert_1 }{\alpha^2} )  , \alpha^4 d^2 \}]$}.
\end{lemma}

\begin{proof}
According to \Cref{app:lemma:B:convergence_iterates}, the flow converges
to an interpolator  $\beta_\infty^\alpha$. We consider the same Lyapunov as before:
\begin{align*}
    W_t &= \phi_{\alpha_\infty}(\beta_\infty^\alpha) - \phi_{\alpha_t}(\beta_{t} )  + 
\langle \nabla \phi_{\alpha_t}(\beta_t) , \beta_t - \beta_\infty^\alpha \rangle,
\end{align*}
which is such that, following the same computations as in \Cref{app:lem:ito_V}:
\begin{align*}
    2 \int_0^t L(\beta_s) \dd s &= W_0 - W_t  +  \gamma  \langle \diagH,  \int_0^t L(\beta_s) \xi_s \dd s \rangle   + S_t \\
    &\geq W_0 - W_t + S_t,
\end{align*}
where $S_t = \int_0^t \sqrt{\gamma L(\beta_s)} \langle X^{\top}   \dd B_s, \beta_s - \betasparse \rangle$.

Now since we put ourselves on  $\mathcal{B}$:
\begin{align*}
    2 \int_0^\infty L(\beta_s) \dd s  
    &\geq W_0 - \hat{a} - 2 \hat{b} \gamma \lambda_{\max} \int_0^{+\infty}  L(\beta_s) (\Vert \beta_s \Vert_1^2 + \Vert \betasparse \Vert_1^2) \dd s  \\
    &\geq W_0 - \hat{a} - 2 \hat{b} \gamma \lambda_{\max} (18^2 + 1 ) \max\Big( \Vert \betasparse \Vert_1^2 \ln^2\Big(\sqrt{2} \frac{ \Vert \betasparse \Vert_1 }{\alpha^2}\Big),   \alpha^4 d^2\Big) \int_0^{+\infty}  L(\beta_s)  \dd s  \\
    &\geq W_0 - \hat{a} -  2 \gamma \hat{b} M \ln(4 / p)^{-1} \int_0^{+\infty}  L(\beta_s)  \dd s,  
\end{align*}
where the second inequality comes from 
\Cref{lem:boundedness} (which is still valid since we are on the event 
$\mathcal{A}$) 
and \mbox{$M = \big [ 325 \ln(4 / p)  \lambda_{\max}   \max( \Vert \betasparse \Vert_1^2 \ln^2 (\sqrt{2} \frac{ \Vert \betasparse \Vert_1 }{\alpha^2}),   \alpha^4 d^2) \big ]$}. 
Hence, we can lowerbound the integral as
\begin{align*}
    \int_0^{+\infty} L(\beta_s) \dd s  
    &\geq \frac{W_0 - \hat{a}}{2 + 2 \gamma \hat{b} M \ln ( \frac{4}{p} )^{-1}}.
\end{align*}
Importantly $W_0 = \phi_{\alpha_{\infty}}(\beta_{\infty}) - \phi_\alpha(0)$ depends on $\beta_\infty$ and is therefore stochastic. 
However, since for all $\beta \in \R^d$, $\alpha \mapsto \phi(\beta, \alpha^2)$ is decreasing and  $\alpha_\infty \leq \alpha$, we obtain:
\begin{align*}
W_0 &= \phi_{ \alpha_{\infty}}(\beta_{\infty}) - \phi_\alpha(0) \\
&\geq \phi_\alpha(\beta_{\infty}) - \phi_\alpha(0) \\
&\geq \phi_\alpha(\beta^*_\alpha) - \phi_\alpha(0):= W_0^{\alpha},
\end{align*}
where $\beta^*_\alpha = \underset{\beta \ \text{s.t} \ X \beta = Y}{\text{argmin}} \phi(\beta, \alpha^2)$.
Therefore, we control the integral of the loss as
\begin{align*}
    \int_0^{+\infty} L(\beta_s) \dd s  
        &\geq \frac{W_0^\alpha - \hat{a}}{2 + 2 \gamma \hat{b} M \ln ( \frac{4}{p} )^{-1}}  \\
    \end{align*}
We now plug in the values $\hat{a} = \frac{W_0^\alpha}{2} $ and $\hat{b} = \frac{1}{ W_0^\alpha} \ln ( \frac{4}{p} ) $:
    \begin{align*}
    \gamma \int_0^{+\infty} L(\beta_s) \dd s  
        &\geq \frac{W_0^\alpha}{4} \frac{\gamma}{1 + \gamma \frac{M}{W_0^\alpha}  }.
    \end{align*}
\end{proof}
To complete our understanding of the dependency of the integral of the loss in terms of $\alpha$ and $\beta^*_{\ell_1}$
we need to know  the dependency of $W_0^\alpha$ in $\alpha$. The following lemma does so. We  consider 
the limit $\alpha \to 0$ which corresponds to the rich regime we are interested in.
\begin{lemma}
\label{app:lem:lowerbound_explicit}
On $\mathcal{A}\cap\mathcal{B}$, let    $\gamma \leq [400 \lambda_{\max}  \ln ( \frac{4}{p} )  \max\{ \Vert \beta_{\ell_1}^{*} \Vert_1 \ln( \frac{\sqrt{2} \Vert \beta_{\ell_1}^{*} \Vert_1 }{\alpha^2 } )  , \alpha^2 d \}]^{-1} $,
then for $\alpha$ small enough:
$$     \int_0^{+\infty} L(\beta_s) \diff s 
\geqslant 
\frac{1}{8}   \Vert \betasparse \Vert_1   \ln \Big (  \frac{ \Vert \betasparse \Vert_1 }{\alpha^2}    \Big )  .
      $$
\end{lemma}
\begin{proof}
Applying Lemma~\ref{app:lemma:LoulouJeTeKiffe}, for all $ \beta \in \R^d\ $,
$\phi_\alpha(\beta) - \phi_\alpha(0) \geq \frac{1}{4} \sum_i  \max \big \{ 0, \vert \beta_i  \vert \ln \frac{\vert \beta_i \vert}{2 \alpha^2} \big \}$.
Therefore,
\begin{align*}
    W_0^\alpha &\geq \frac{1}{4} \sum_{i}  \vert \beta_{\alpha, i}^* \vert \ln \frac{\vert \beta_{\alpha, i}^*  \vert}{2 \alpha^2} .
\end{align*}
Note that 
$\beta^*_\alpha = \underset{\beta \ \text{s.t} \ X \beta = Y}{\text{argmin}} \phi_\alpha(\beta)$
and 
$\beta^*_{\ell_1} = \underset{\beta \ \text{s.t} \ X \beta = Y}{\text{argmin}} \Vert \beta \Vert_1$.
From Theorem 2 of \cite{woodworth2020kernel}:
~$ \Vert \beta^*_\alpha \Vert_1 \underset{\alpha \to 0}{\longrightarrow}   \Vert \beta_{\ell_1}^* \Vert_1 $ which leads to:
\begin{align*}
 \sum_{i}  \vert \beta_{\alpha, i}^* \vert \ln \frac{\vert \beta_{\alpha, i}^*  \vert}{2 \alpha^2} 
    & \underset{\alpha \to 0}{\sim} \Vert \beta_{\ell_1}^* \Vert_1  \ln \frac{\Vert \beta_{\ell_1}^* \Vert_1}{\alpha^2} .
\end{align*}
and 
$  W_0^\alpha \underset{\alpha \to 0}{\geqslant}   \frac{1}{4}   \Vert \betasparse \Vert_1   \ln \big (  \frac{ \Vert \betasparse \Vert_1 }{\alpha^2}    \big )  $.
Finally, for $\alpha$ small enough, from the upperbound on $\gamma$, the value of $M$ and the lower bound on $W_0^\alpha$:
\begin{align*}
    \gamma  \frac{M}{W_0^\alpha} \underset{\alpha \to 0}{\leq} 1,
\end{align*}
which along with \Cref{app:lemma:B:gamma_int_loss} concludes the proof.
\end{proof}

Therefore through this lemma we see that by picking the biggest step-size which ensures convergence, we have a dependency
of the integral of the loss as $\ln \frac{1}{ \alpha}$.

Now we are equipped to prove \Cref{prop:loss_integral}. We recall it here to be self-contained.

\lossintegral*

\begin{proof}
Let us place ourselves on the event $\mathcal{A}\cap\mathcal{B}$. Let us recall that $\P(\mathcal{A}\cap\mathcal{B}) \geq 1 - (\P( \mathcal{A}^C) + \P( \mathcal{B}^C)) \geq 1-p$, 
where the last inequality results from the definitions of $\mathcal{A}$ and $\mathcal{B}$ and \Cref{app:lemma:martingale_result}. As this event is included in $\mathcal{A}$, the right inequality of the proof corresponds exactly to the \Cref{app:prop:loss_integral} of \Cref{app:subsec:upperbound_loss}. The proof of left inequality of the proposition comes from \Cref{app:lem:lowerbound_explicit}.
\end{proof}

In the final proposition of this subsection, we give the scale of $\alpha_{\infty}$ we obtain thanks to our analysis.  
Indeed though we know that in all case $\alpha_\infty < \alpha$, we would like to quantitatively 
know
\textbf{how much} smaller the effective initialisation is in order to have an idea of
the gain of  SGD over GD (in terms of implicit bias).

\begin{prop}
    \label{app:prop:scale_alpha_eff}
Consider the iterates $(w_t)_{t \geq 0}$ issued from the stochastic gradient flow \eqref{eq:SGF}, initialised at $w_{0, \pm} = \alpha \mathbf{1} \in (\R_{+}^{*})^d$.  
Let $p \leq \frac{1}{2}$ and $\gamma$ matching the upperbound in \Cref{thm:main_theorem}, i.e. 
$\gamma = [400 \lambda_{\max}  \ln ( \frac{4}{p} )  \max\{ \Vert \beta_{\ell_1}^{*} \Vert_1 \ln( \frac{\sqrt{2} \Vert \beta_{\ell_1}^{*} \Vert_1 }{\alpha^2 d} )  , \alpha^2 d\}]^{-1} $,
then
with probability at least $1 - p$ and for $\alpha$ small enough:
\begin{align*}
    \frac{\alpha_\infty}{\alpha} \leq  \exp \Big (  - \frac{1}{1600  \ln(\frac{4}{p}) } \frac{\diag( \frac{X^\top X}{n}) }{\lambda_{\mathrm{max}} } \Big ).
\end{align*}

\end{prop}

\begin{proof}
The fact that 
$\alpha_{\eff} = \alpha  \exp\left( - 2 \gamma \diag\left( \frac{X^{\top} X}{n} \right) \int_0^{+ \infty}  L(\beta_s) \diff s \right)$
along with the lower bound from
    \Cref{app:lem:lowerbound_explicit} and the value of $\gamma$
    gives the result.
\end{proof}

This result tends to show that the overall gain of SGD over GD is only by 
a constant factor 
$\exp (  - \frac{1}{1600  \ln(\frac{4}{p}) } \frac{\diag( \frac{X^\top X}{n}) }{\lambda_{\mathrm{max}} }) < 1$.
We believe that our analysis is not tight and that the gain is in fact more consequent, 
this is explained in the following subsection.

\subsection{Scale of \texorpdfstring{$\alpha_\infty$}{alphainfty} when assuming that the iterates are bounded independently of \texorpdfstring{$\alpha$}{alphainfty}.}
\label{app:B:true_scale_alpha_inf}

In this subsection we explain why we believe that our analysis lacks of tightness. 
In \Cref{lem:boundedness} there is a dependency in $\ln(\frac{1}{\alpha})$ 
in the upperbound of the $\ell_1$ norm of the iterates. We believe that this dependency is
an artifact of our analysis and that the true bound is independent of $\alpha$, this is also
what is observed in practice. This is the reason why we formulate the following assumption:

\begin{assumption*} 
    \label{app:boundedness_assumption}
    On $\mathcal{A}$, $ \Vert \beta_t \Vert_1 \leq \Vert \xi_t \Vert_1  \leq \max \{ \Vert \beta_{\ell_1}^* \Vert_1 ,   \alpha^2 d \}$ for all $t \geq 0$.
\end{assumption*}

 Under this assumption, we obtain convergence of the iterates towards 
 an interpolating solution under a weaker constraint on $\gamma$ (bigger step-sizes can be used while still ensuring convergence)
 as well as a much better upperbound on the scale of $\alpha_\infty$. The aim of the following result is to give the relevant scale of how small is $\alpha_\infty$ w.r.t. $\alpha$. Hence, for the sake of clarity, we will assume that $\diag (X^\top X / n) \sim \lambda_{\mathrm{max}} \mathbf{1}$ (which is true for sub-gaussian inputs with high probability). We also fix $p=0.01$ and drop all the numerical constants under some universal constant $\zeta>0$.

 \begin{prop}
    \label{app:prop:true_scale_alpha_eff}
Consider the iterates $(w_t)_{t \geq 0}$ issued from the stochastic gradient flow \eqref{eq:SGF}, initialised at $w_{0, \pm} = \alpha \mathbf{1} \in (\R_{+}^{*})^d$.  
Assume boundedness of the iterates and  $\gamma = \Theta \big ( \max\{ \Vert \beta_{\ell_1}^{*} \Vert_1  , \alpha^2 d \}^{-1} \big ) $,
then
with probability at least $0.99$, the iterates $(\beta_t)_{t \geq 0}$ converge towards an interpolating solution 
$\beta_\infty^\alpha = 
\underset{\beta \in \R^d \ \text{s.t.} \ X \beta = y}{ \mathrm{arg \ min \ }} \phi_{\alpha_{\eff}}(\beta) $. Furthermore, for $\alpha$ small enough, there exists $\zeta>0$ such that:
\begin{align*}
    \frac{\alpha_\infty}{\alpha} \leq  \left(  \frac{\alpha^2}{\Vert \beta_{\ell_1}^* \Vert_1 } \right)^\zeta.
\end{align*}

\end{prop}

\begin{proof}
As said earlier, we fix $p=0.01$. Then, by following the proof of 
\Cref{app:lem:control_weight}, and using the boundedness assumption 
instead of \Cref{lem:boundedness}, one obtains that for 
$\gamma \leq O \big ( \max\{ \Vert \beta_{\ell_1}^{*} \Vert_1  , \alpha^2 d \}^{-1} \big ) $ 
(as mentioned the precise numerical constants are 
dropped for simplicity) then $U_t \geq \frac{1}{2}$ for all $t \geq 0$.
The results of \Cref{app:B:lemma:lower_bound_Vt}, \Cref{app:prop:loss_integral}, 
\Cref{app:lemma:B:convergence_iterates} and 
therefore \Cref{thm:main_theorem}
then still hold with probability $0.99$ but with 
the weaker condition that 
$\gamma \leq O \big ( \max\{ \Vert \beta_{\ell_1}^{*} \Vert_1  , \alpha^2 d \}^{-1} \big ) $. 

For the upperbound on $\alpha_\infty$, we follow the exact same steps as in 
\Cref{app:subsec:lowerbound_loss}. Indeed 
\Cref{app:lemma:B:gamma_int_loss} now gives, for $\gamma \leq O \big ( (\lambda_{\mathrm{max}} \max\{ \Vert \beta_{\ell_1}^{*} \Vert_1  , \alpha^2 d \})^{-1} \big )$:
\begin{align*}
    \gamma \int_0^{+ \infty} L(\beta_s) \dd s 
    &\geq \frac{W_0^\alpha}{4} \frac{\gamma }{1 + \gamma \frac{M}{W_0^\alpha}  },
\end{align*}
where
$M = \Theta \big (  \lambda_{\mathrm{max}} \max\{ \Vert \beta_{\ell_1}^{*} \Vert_1^2  , \alpha^4 d^2 \} \big )$.
Plugging in the maximum value of $\gamma$, i.e. 
$\gamma = \Theta \big ( (\lambda_{\mathrm{max}} \max\{ \Vert \beta_{\ell_1}^{*} \Vert_1  , \alpha^2 d \})^{-1} \big )$:  we have that 
$\gamma \frac{M}{W_0^\alpha} \underset{\alpha \to 0}{\longrightarrow} 0$ and for $\alpha$ small enough
$\gamma  W_0^\alpha \geq \Omega \left ( \lambda_{\mathrm{max}}^{-1} \ln \left(  \frac{ \Vert \betasparse \Vert_1 }{\alpha^2}  \right)     \right ) $.
Therefore for $\alpha$ small enough:
\begin{align*}
    \gamma \int_0^{+ \infty} L(\beta_s) \dd s 
    &\geq \Omega \left( \lambda_{\mathrm{max}}^{-1} \ln \left (  \frac{ \Vert \betasparse \Vert_1 }{\alpha^2}  \right)\right)
\end{align*}
Plugging this inequality into the definition of $\alpha_\eff$ and assuming that 
$\diag (X^\top X / n) \sim \lambda_{\mathrm{max}} \mathbf{1}$ leads to:
\begin{align*}
    \alpha_\infty = \alpha \exp\left(\!- 2 \diag\left( \frac{X^{\top} X}{n} \right)\gamma\!\int_0^{+ \infty}\!\!\!\!\!\!\!\! L(\beta_s) \diff s \!\right) \leq  \alpha \left(  \frac{\alpha^2}{\Vert \beta_{\ell_1}^* \Vert_1 } \right )^{ \Omega \big ( 1 \big )}.
\end{align*}
This concludes the proof of the Proposition.
\end{proof}
This upperbound is significantly better than that of 
\Cref{app:prop:scale_alpha_eff}: the smaller the initialisation scale $\alpha$ and the greater the 
benefit  of SGD over GD in terms of implicit bias. More precisely, Proposition \ref{app:prop:true_scale_alpha_eff} shows that the benefit scales as a power law with respect to the initialization $\alpha$.






    
\label{app:proofs}


\section{Deterministic framework}
\label{app:sec:deterministic}

In this section we recall some known results concerning the implicit bias of deterministic mirror descent 
as well as give convergence guarantees.
%
In the previous section, the stochasticity of the flow made the analysis much more involved. 
In contrast, the analysis is straightforward in the deterministic setting and we believe this simple case can serve as a warmup to gain further intuition. 
 %
Note that even though these results are known independently, we did not find a clear reference gathering them.
See for example \cite{bauschke2017descent} for the convergence of the iterates towards an interpolator and 
\cite{gunasekar2018characterizing} for the associated implicit minimisation problem.

%
\begin{prop}
\label{app:prop:mirror_descent}
For any convex loss $L$ such that there exists at least one zero-error interpolator, let $\Psi: \R^d \to \R$ be a strongly convex and twice differentiable function which we call potential. 
For any initialisation $\beta_0 \in \R^d$, consider the mirror descent flow  $(\beta_t)_t$: 
\begin{align}
\label{app:ode:mirror_descent}
\diff \nabla \Psi(\beta_t) =  - \nabla L(\beta_t) \mathrm{d} t.
\end{align}
Then the iterates $(\beta_t)_t$ converge to an interpolator $\beta_\infty$ which satisfies:
\begin{align}
\label{app:minimisation_problem_GF}
    \beta_{\infty} = \underset{\beta \in \R^d}{ \argmin } \ D_{\Psi}(\beta, \beta_0) \quad \text{such that } \quad X \beta = y,
\end{align}
where $D_{\Psi}(\beta, \beta_0) = \Psi(\beta) - \Psi(\beta_0) - \langle \nabla \Psi(\beta_0), \beta - \beta_0 \rangle$ is the Bregman divergence w.r.t. $\Psi$.
\end{prop}

\begin{proof} We  divide the proof into three steps.

\textbf{First step: the loss goes to $0$.} 

Note that:
\begin{align*}
  \frac{\diff}{\diff t} L(\beta_t) &=  - \langle \nabla L(\beta_t), \dot{\beta}_t     \rangle \\
  &=  - \langle [\nabla^2 \Psi(\beta_t)]^{-1} \nabla L(\beta_t) , \nabla L(\beta_t)     \rangle \\
  &\leq 0,
\end{align*}
where the inequality is by convexity of the potential $\Psi$.
Hence the loss is decreasing.
Now consider the Bregman divergence between an arbitrary interpolator $\beta^*$ and $\beta_t$:
\begin{align*}
  D_{\Psi}(\beta^*, \beta_t ) = \Psi(\beta^*) - \Psi(\beta_t) - \langle \nabla \Psi(\beta_t), \beta^* - \beta_t \rangle \geq 0.
\end{align*}
which is such that:
\begin{align}
  \label{ode:mirror_descent_proof}
  \frac{\diff}{\diff t }  D_{\Psi}(\beta^*, \beta_t ) &=  \langle \frac{\diff}{\diff t } \nabla \Psi(\beta_t), \beta_t - \beta^* \rangle \nonumber \\ 
        &= - \langle  \nabla L(\beta_t), \beta_t - \beta^* \rangle  \nonumber \\
        &\leq - L(\beta_t) \\
        &\leq 0 \nonumber
\end{align}
where the first inequality is by the convexity of the loss. Therefore:
\begin{align*}
L(\beta_t) 
  &\leq   \frac{1}{t} \int_0^t L(\beta_s) \diff s  \\ 
  &\leq  \frac{ D_\Psi(\beta^*, \beta_0) - D_\Psi(\beta^*, \beta_t)   }{t} \\
  &\leq  \frac{ D_\Psi(\beta^*, \beta_0)    }{t} \\
   &\underset{t \to +\infty}{\longrightarrow} 0.
\end{align*}
Hence the loss converges to $0$. 

\textbf{Second step: the iterates converge towards an interpolator $\beta_\infty$.}

Since 
$
  \frac{\diff}{\diff t }  D_{\Psi}(\beta^*, \beta_t ) 
        \leq 0 $,
we have that whatever the interpolator $\beta^*$,  $D_{\Psi}(\beta^*, \beta_t )$ is decreasing over the trajectory. Since it is a positive 
quantity we get that it converges. 
Moreover $\Psi$ is $\mu$-strongly convex which means that we also have $\Vert \beta_t - \beta^* \Vert_2^2 \leq \frac{2}{\mu} D_{\Psi}(\beta^*, \beta_t )$. 
The flow $(\beta_t)_t$ is therefore bounded and we can extract a convergent subsequence: let $\beta_\infty$
be such that $\beta_{\phi(t)} \underset{t \to \infty}{\longrightarrow} \beta_\infty$. 
Since from the first step $L(\beta_t) \to 0$, by unicity of the limit $L(\beta_{\phi(t)})$ also converges to $0$, and we get by continuity of $L$  that $\beta_\infty$ is an interpolator.
This means that   (a)  $D_{\Psi}(\beta_\infty, \beta_t )$ converges and (b) it  converges towards the same limit as $D_{\Psi}(\beta_\infty, \beta_{\phi(t)} )$ which is $0$.
Finally:
\begin{align*}
  0\leq \Vert \beta_t - \beta_{\infty} \Vert_2^2 \leq \frac{2}{\mu} D_{\Psi}(\beta_\infty, \beta_t )  \underset{t \to \infty}{\longrightarrow} 0,
\end{align*} 
and therefore $\beta_t$ converges towards the interpolator $\beta_\infty$.

\textbf{Third step: implicit bias.}

Note that:
\begin{align*}
  \nabla \Psi(\beta_t) - \nabla \Psi(\beta_0) 
  &= - \int_0^t \nabla L(\beta_s) \diff s \\
  &= - X^{\top} \int_0^t (X \beta_s - y) \diff s \in \mathrm{span}(X).
\end{align*}
Therefore: $\nabla \Psi(\beta_\infty) - \nabla \Psi(\beta_0) \in \mathrm{span}(X)$ and $X \beta_\infty = y$, which are exactly the 
KKT conditions of the minimisation problem: 
\begin{align*}
 \underset{\beta \in \R^d}{ \min } \ D_{\Psi}(\beta, \beta_0) \quad \text{such that } \quad X \beta = y.
 \end{align*}

\textbf{Remark on the loss integral.} We can also show that the integral of the loss converges.
Indeed from inequality \ref{ode:mirror_descent_proof} with $\beta^* = \beta_\infty$, we immediately get that 
$\int_0^\infty L(\beta_s) \dd s \leq D_\Psi(\beta_\infty, \beta_0)$. Furthermore, when $L$ is the square loss 
$L(\beta) = \frac{1}{2} (\beta - \beta^*)^\top H (\beta - \beta^*)$, then inequality \ref{ode:mirror_descent_proof} 
becomes the equality $ \frac{\diff}{\diff t }  D_{\Psi}(\beta^*, \beta_t ) = - 2 L(\beta_t)$ and hence
$\int_0^\infty L(\beta_s) \dd s = \frac{1}{2} D_\Psi(\beta_\infty, \beta_0)$.

%
%
%
\end{proof}
In our framework, for the deterministic case, we cannot simply apply this result 
with $\phi_\alpha$, indeed it is not strongly convex over $\R^d$.
However following the exact same proof as in \Cref{lem:boundedness}  and \Cref{app:lem:control_weight}
but in the deterministic case (which is easier since we do not need 
to use martingale concentration inequalities), we can show that the iterates $\beta_t$ are bounded. 
Using \Cref{app:prop:mirror_descent} but on a convex set in which the iterates 
stay and over which $\phi_\alpha$ is strongly convex (as done in 
\Cref{app:lemma:B:convergence_iterates} ) leads to  the convergence of the iterates.

\label{app:mirror_deterministic}


\section{Experiments}
\label{app:section:experiments}

In the following section we consider the same experimental setup as in \Cref{subsection:experim_setup}, which we recall here for clarity.
We consider $n = 40$, $d = 100$ and randomly 
generate a sparse model $\beta^*_{\ell_0}$ such that $\Vert \beta^*_{\ell_0}  \Vert_0 = 5$. We generate the features as $x_i \sim \mathcal{N}(0, I)$
and the labels as $y_i = x_i^{\top} \beta^*_{\ell_0}$. 
We use the same step size for GD and SGD and choose it to be the biggest as possible while still ensuring convergence. Note
that since the true population covariance $\mathbb{E}[x x^{\top}] $ is equal to identity, the quantity  $\Vert \beta_t - \beta^*_{\ell_0} \Vert_2^2$ corresponds to the validation 
loss.

\subsection{Doping the implicit bias using label noise: experiments}
\label{app:subsec:label_noise}

We consider the label noise setting discussed in \Cref{main:subsec:label_noise}: 
for a sequence $(\delta_t)_{t \in \mathbb{N}} \in \R_+$, assume 
that we artificially inject some label noise $\Delta_t$ at time $t$, say for example
$\Delta_t \sim \mathrm{unif} \{2 \delta_t, -2 \delta_t \} $ and independently from $i_t$ (other type 
of label noise can of course be considered, but we consider here this one for simplicity). 
This injected label noise perturbs the SGD recursion as follows:
\begin{align}
 w_{t+1, \pm} = w_{t, \pm} \mp \gamma  \left(\langle \beta_w - \beta^*, x_{i_t} \rangle + \Delta_t\right) \  x_{i_t} \odot w_{t, +}\ ,
 \qquad \  \text{where }\  i_t \sim \mathrm{unif}(1, n).
\end{align}
Using the same notations and following the same derivations as in 
\Cref{app:subsec:details_SDE_model}, we can rewrite the recursion as:
\begin{align*}
   w_{t+1, \pm} 
      &= w_{t, \pm} - \gamma  \nabla_{w_{\pm}} L(w_t) \pm \gamma \diag(w_{t, \pm}) X^{\top} [ \xi_{i_t}(\beta_t) + \Delta_t \mathbf{e}_{i_t} ].
 \end{align*}
Since $\Delta_t$ is zero-mean and independent of $i_t$ we get:
\begin{align*}
   \text{Cov}_{i_t}[ \xi_{i_t}(\beta) + \Delta_t \mathbf{e}_{i_t}] &= \mathbb{E}_{i_t}[ \xi_{i_t}(\beta)^{\otimes 2} ] + \mathbb{E}[ \Delta_t^2 \mathbf{e}_{i_t}^{\otimes 2} ] \\
   &= \mathbb{E}_{i_t}[ \xi_{i_t}(\beta)^{\otimes 2} ] +  \frac{4 \delta_t^2}{n} I_n.
\end{align*}
Now following the same reasoning as in \Cref{app:subsec:details_SDE_model}, 
it is natural to consider the following SDE: 
\begin{align*}
  \dd w_{t,\pm} &= - \nabla_{w_\pm} L(w_t) \dd t \pm 2 \sqrt{\gamma n^{-1} (L (w_t) + \delta_t^2)}\  w_{t,+} \odot [X^\top \dd  B_t].
 \end{align*}
Let $\tilde{L}(\beta_t) =  L(\beta_t) + \delta_t^2$ be the "slowed down" loss.
Following the same computations as for \Cref{app_lemma:closed_form_beta}
we obtain that:
\begin{align*}
       \beta_t &= 2 \tilde{\alpha}_t^2  \odot \sinh ( 2 \tX^{\top} \tilde{\eta}_t ) ,
   \end{align*}
where 
$\tilde{\eta}_t =  - \int_0^t  \tX (\beta_s - \beta^*) \diff s + 2 \sqrt{\tgamma } \int_0^t \sqrt{\tilde{L}(\beta_s)}  \dd B_s \in \R^n$
and
$\alpha_t = \alpha \odot  \exp \big  ( - 2 \tgamma \diag(\tX^{\top} \tX) \int_0^t \tilde{L}(\beta_s) \dd s \big )$.
And following the proof of \Cref{prop:sto_mirror_descent}:
\begin{align}
       \diff \nabla \phi_{\alpha_t}(\beta_t) = 
        - \nabla L(\beta_t) \diff t + \sqrt{\gamma n^{-1} \tilde{L}(\beta_t)} X^{\top} \mathrm{dB}_t.
   \end{align}
Assuming that $(\delta_t)_{t \geq 0} \in (\R_+)^\R$ and $\gamma$ are such that the iterates  converge (here we do not  
show under which conditions we have convergence and leave this as future work), 
the corresponding implicit regularisation minimisation problem is preserved 
but with an effective initialisation: 
$ \tilde{\alpha}_{\eff} = \alpha \odot \exp{\left(- 2 \gamma \diag(\frac{X^\top X}{n}) \int_0^{+ \infty} \tilde{L}(\beta_s)  \diff s\right)}$
which takes into account the slowed down loss 
$\tilde{L}(\beta_t) =  L(\beta_t) + \delta_t^2$ .
Since it is reasonable to consider that $\tilde{\alpha}_\infty < \alpha_{\infty}$,  
the label noise therefore helps to recover a solution which has better 
sparsity properties.

\begin{figure}[h]
    \centering
    \begin{minipage}[c]{.32\linewidth}
    \hspace*{-15pt}
    \includegraphics[width=\linewidth]{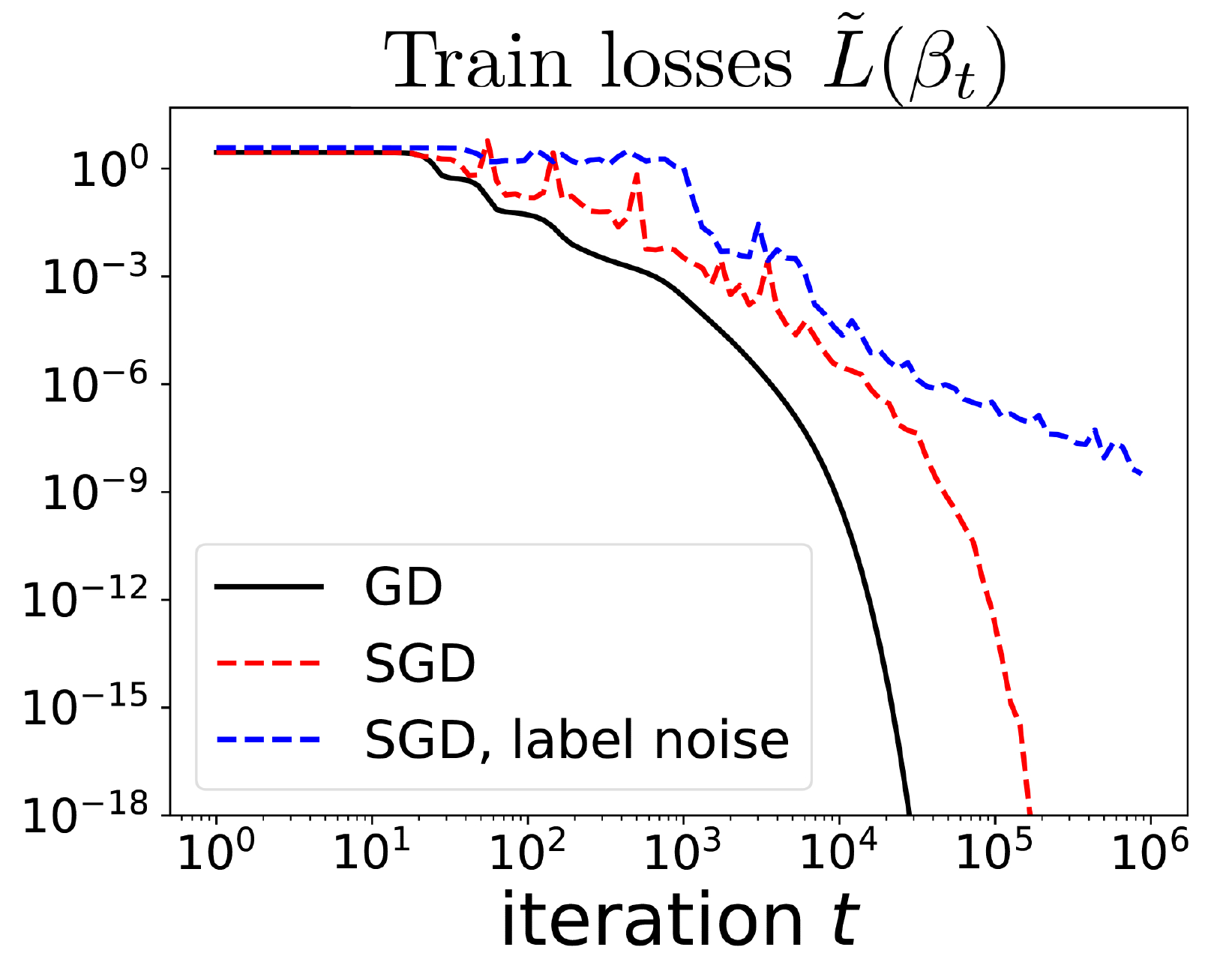}
       \end{minipage}
       \hspace*{-5pt}
       \begin{minipage}[c]{.32\linewidth}
    \includegraphics[width=\linewidth]{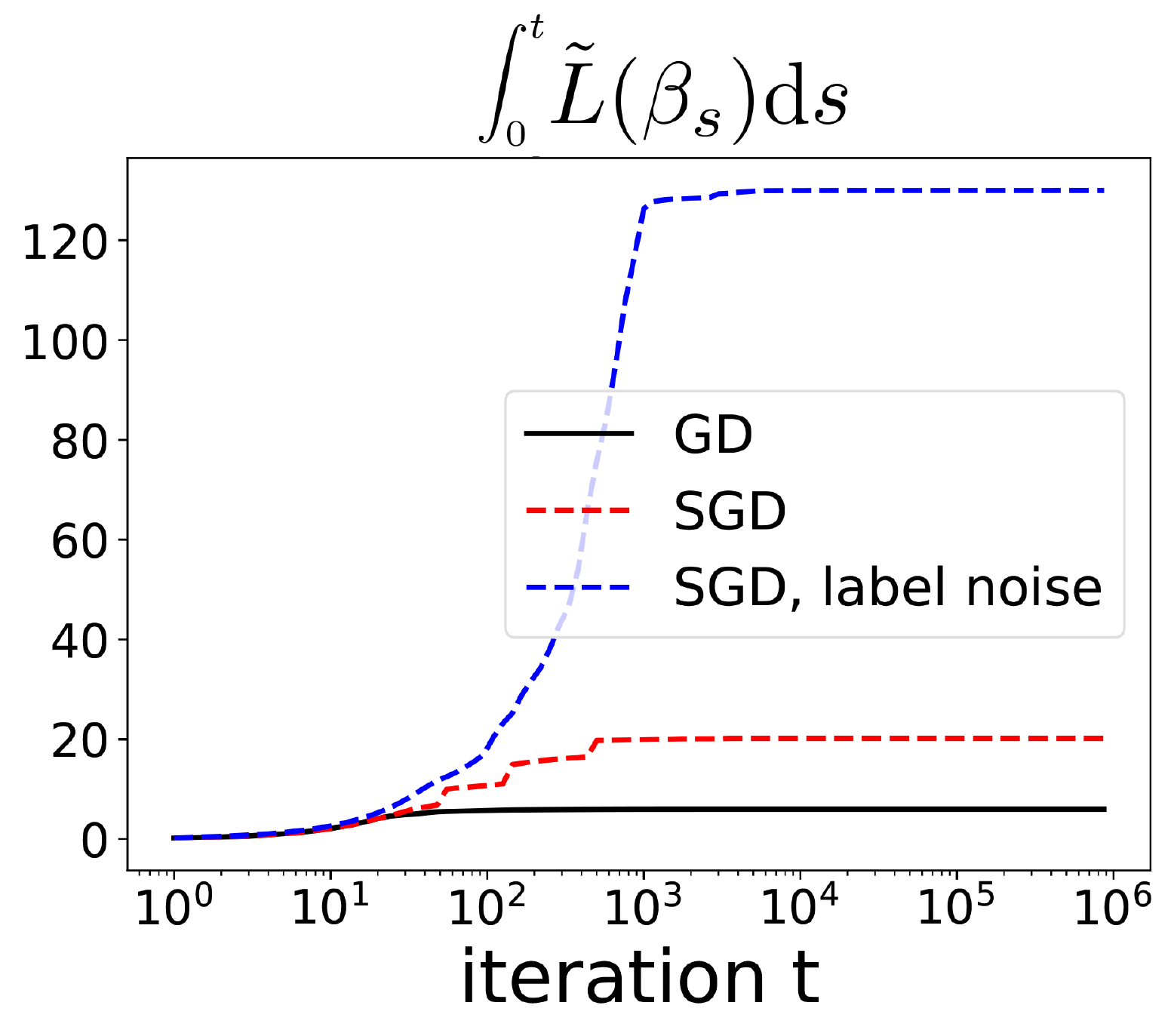}
       \end{minipage}
        \hspace*{0pt}
       \begin{minipage}[c]{.32\linewidth}
    \includegraphics[width=\linewidth]{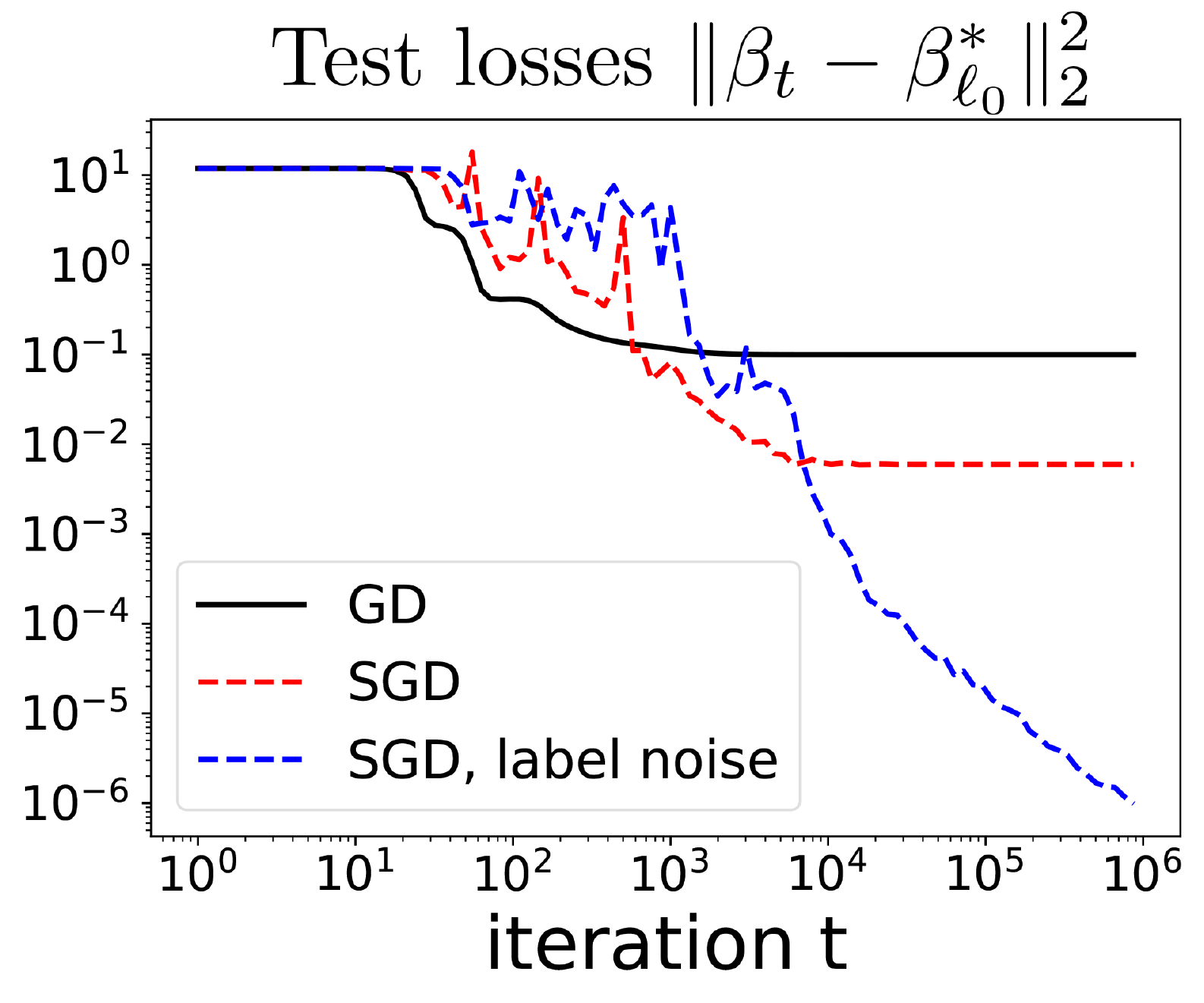}
       \end{minipage}
       \caption{ Sparse regression (see \Cref{subsection:experim_setup} for the detailed experimental setting), illustration of the 
       benefits of using label noise. 
           All experiments are initialised at $\alpha = 0.01$.
        \textit{Left}: The use of label noise slows down the convergence of the effective training loss $\tilde{L}$.
        \textit{Middle and right}: the value of the integral of the slowed down loss $\tilde{L}$ is much higher for the recursion 
        with label noise, leading to a solution which generalises much better.
         }
         \label{app:fig:label_noise}
    \end{figure}

We experimentally validate the advantage of adding label noise by choosing 
the sequence $\delta_t = 1$ if $t \leq 10^3$ and $\delta_t = 0$ if $t > 10^3$.
The results are illustrated \Cref{app:fig:label_noise}. Note that the training loss is heavily slowed down, however the recovered 
solution at iteration $t = 10^6$ is much better than that of SGD, and it has not even converged yet. 
However, it must be kept in mind that 
adding too much label noise can significantly slow down the convergence of the 
validation loss or even
prevent the iterates from converging.

\label{app:experiments}


\section{Extensions}

We introduce two extensions of our results: subsection~\ref{app:subsection:E:general_SDE_model} extends our results for a very general stochastic gradient flow model  and subsection~\ref{app:subsec:depth_p} discuss them in the depth $p\geq 3$ case.

\subsection{Towards a more general SDE modelling}
\label{app:subsection:E:general_SDE_model}

Recall from the SDE modelling of  Appendix~\ref{app:SDE_modelling}  that 
$  \text{Cov}_{i_t}[ \xi_{i_t}(\beta) ] = \frac{4}{n} \diag (L_i(\beta) )_{1 \leq i \leq n} + O(\frac{1}{n^2})$.
If we assume $n$ large enough we can neglect the second order term of order $1/n^2$:
\begin{align*}
     \text{Cov}_{i_t}[ \xi_{i_t}(\beta) ] 
     &\cong \frac{4}{n} \diag (L_i(\beta) )_{1 \leq i \leq n}.
\end{align*}
Assume we do not consider that $L_i(\beta) \sim L(\beta)$, then the 
 overall SGD noise structure is captured by 
\begin{align*}
     \Sigma_{_{\mathrm{SGD}}} (w_\pm) 
     &:= \gamma^2 \diag(w_\pm) X^{\top} \text{Cov}_{i_t}[ \xi_{i_t}(\beta) ] X \diag(w_\pm) \\
&\cong \frac{4}{n} \gamma^2 [\diag(w_\pm) X^{\top} \diag (\sqrt{L_i(\beta) })]^{\otimes 2}.
\end{align*}

This leads us in considering the following SDE:
\begin{equation}
    \begin{aligned}
    \dd w_{t,+} &= - \nabla_{w_+} L(w_t) \diff t + 2 \sqrt{\gamma  } \  w_{t,+} \odot [\tX^\top \diag (\sqrt{L_i(\beta) }) \dd  B_t]  \\
    \dd w_{t,-} &= - \nabla_{w_-} L(w_t) \diff t - 2 \sqrt{\gamma  } \  w_{t,-} \odot [\tX^\top \diag (\sqrt{L_i(\beta) }) \dd  B_t].
    \end{aligned}
    \end{equation}
As previously, this SDE admits an implicit integral formulation (multiplication must be understood component-wise):
\begin{align*}
    w_{t, \pm}  & = w_{t = 0, \pm} \odot \exp ( \pm \tX^{\top} \big [ - \int_0^t r(w_s) \diff s + 2 \sqrt{ \tgamma } \int_0^t \diag (\sqrt{L_i(w_s) }) \diff B_s \big ]  )    \\
    &\quad \quad \odot \exp ( - 2 \tgamma  \diag(\tX^{\top } \int_0^t  \diag (L_i(w_s) ) \diff s  \tX)    ) \\
     & = \alpha_t \odot \exp ( \pm \tX^{\top} \eta_t )   , 
\end{align*}
where 
$\eta_t =  - \int_0^t  \tX (\beta_s - \beta^*) \diff s + 2 \sqrt{ \tgamma } \int_0^t \diag (\sqrt{L_i(w_s) }) \diff B_s \in \R^n$
and
$\alpha_t = \alpha \odot \exp ( - 2 \tgamma \diag(\tX^{\top } \int_0^t  \diag (L_i(w_s) ) \diff s  \tX)    )$.
Since $\beta = w_+^2 - w_-^2$, we get:
\begin{align*}
    \beta_t &= \alpha_t^2 \odot \big ( \exp ( + 2 \tX^{\top} \eta_t  ) - \exp ( - 2 \tX^{\top} \eta_t  ) \big ) \\
            &= 2 \alpha_t^2 \odot \sinh ( + 2 \tX^{\top} \eta_t  ). 
\end{align*}
And we obtain the following mirror-type descent flow:
\begin{align*}
    \diff \nabla \phi_{\alpha_t}(\beta_t) = 
        - \nabla L(\beta_t) \diff t +  \sqrt{ \tgamma } \tX^{\top} \diag (\sqrt{L_i(\beta_t) }) \diff B_t.
\end{align*}

\textbf{Assuming convergence of the iterates and of $\alpha_t$} (we do not show the convergence, 
though we think the proof could straightforwardly be adapted following 
\Cref{app:section:B}
 ), the corresponding minimisation problem is:
\begin{equation*}
 \beta_\infty^\alpha = \underset{\beta \in \R^d \ \text{s.t.} \ X \beta = y}{ \mathrm{arg \ min \ }} \phi_{\alpha_{\eff}}(\beta) \quad 
 \text{where } \ \ \alpha_{\eff} = \alpha \odot \exp ( - 2 \tgamma \diag(\tX^{\top } \int_0^\infty  \diag (L_i(\beta_s) ) \diff s \ \tX)    ).
\end{equation*}
Note that the main result of the  paper is very similar,
the difference relies in:
\begin{itemize}
\item  the $k^{th}$ coordinate of
$ \diag(\tX^{\top } \diag (L_i(\beta_s) )  \ \tX) $ is 
$\mathbb{E}_{i_t}[ L_{i_t}(\beta_s) (x_{i_t}^{(k)})^2 ] $ 
\item the $k^{th}$ coordinate of 
$L(\beta_s) \diag(\tX^{\top } \tX) $ is  $\mathbb{E}_{i_t}[ L_{i_t}(\beta_s) ] \mathbb{E}_{i_t}[ (x_{i_t}^{(k)})^2 ] $
\end{itemize}


\subsection{Higher order models: the cases of depth \texorpdfstring{$p > 2 $}{p}}
\label{app:subsec:depth_p}

Until now, we have focused on a $2$-homogeneous parametrisation of the estimator. 
A legitimate question is how the implicit bias changes as we go to a higher degree of homogeneity. 
In terms of networks architecture, this corresponds to increasing the depth of the neural networks. 
Let us fix $p \geqslant 3$ with the new parametrisation $\beta_w = w_+^p - w_-^p$, the loss of our new model writes: $L(w) = \frac{1}{4 n} \sum_{i = 1}^n \langle w_+^p - w_-^p - \beta^*, x_i \rangle^2$.
As previously, we want to consider the stochastic differential equation related to stochastic gradient descent on the above loss. With the same modelling as in Section~\ref{subsec:SDE_model}, stochastic gradient flow writes:
\begin{align}
\dd w_{t,\pm} &= - \nabla_{w_\pm} L(w_t) \dd t \pm 2 \sqrt{\gamma n^{-1} L(\beta_t)} \diag( w^{p-1}_{t,\pm}) X^\top \dd B_t,
\end{align}
where $B_t$ is a standard Brownian motion in $\R^n$. We would like to put emphasis that, unlike the $2$-depth model, we do not provide a dynamical analysis enabling convergence proof and control of interesting quantities. Here, the aim is to show how our framework naturally extends to general depth and how the convergence speed of the loss still seems to controls the effect of the stochastic flow biasing. Contrary to the $2$-depth case, the potential cannot be defined in close form, but we still have the following explicit expression,
$
\phi^p_{ \alpha, \pm}(\beta) = \sum_{i=1}^d \psi^p_{ \alpha, \pm}(\beta_i),
$
where $\psi^p_{ \alpha, \pm} = \int [h^p_{ \alpha, \pm}]^{-1}$ is a primitive of the unique inverse of $h^p_{ \alpha, \pm}(z) := (\alpha_{+}^{2-p} - z)^{-\frac{p}{p-2}} - (\alpha_{-}^{2-p} + z)^{-\frac{p}{p-2}}$ in $(-\alpha_-^{2-p},\alpha_+^{2-p})$.
In the following theorem we characterize the implicit bias of the stochastic gradient flow when applied with higher order models. 
\begin{thm*}
Initialise the stochastic gradient flow with $w_0 = \alpha \mathbf{1} \in \R^{2d}$. If we assume that the flow $(\beta_t)_{t \geq 0}$ converges almost surely towards a zero-training error solution $\beta_\infty^{\alpha, p}$, and that the quantities $\int_0^\infty  L(\beta_s) w_{s,\pm}^{p-2}  \dd s$ and $\int_0^\infty  L(\beta_s)  \dd s$ exist a.s., then the limit satisfies $$\beta_{\infty,p} = \underset{\beta\ s.t\ X \beta = y}{ \mathrm{arg \ min \ }} \phi^p_{\alpha_{\eff}, \pm}(\beta),$$
with ${ \alpha_{\eff, \pm} = \alpha (1 + 2 \gamma (p-2)(p-1) \alpha^{{p-2}} \diag (\frac{X^\top X}{n}) \odot \int_0^\infty  L(\beta_s)  w_{s,\pm}^{p-2}  \dd s)^{^{\!-\frac{1}{p-2}}}} $.
\end{thm*}
First let us stress that without a close form expression of $\phi^d_\alpha$ and proper control of $\int_0^\infty  L(\beta_s) w_{s,\pm}^{p-2}  \dd s$ with respect to $p$ or $\alpha$, it is difficult to conclude directly on the magnitude of the stochastic bias. Yet, the main aspect we can comment on is that, as in the depth-$2$ case, $\alpha_{\eff,\pm} \leqslant \alpha$ almost surely\footnote{Note that, as the weights are initialized positively, they remain positive: $w_{t,\pm}>0$,  for all $t\geqslant0$.} and that the convergence speed of the loss controls the biasing effect. As in \cite{woodworth2020kernel}, it can be shown empirically that $\phi^p_{\alpha,\pm}$ interpolate between the $\ell_1$ and the $\ell_2$ norm as $\alpha_\pm \to 0$ and $\alpha_\pm \to +\infty$ respectively and that the transition is faster than for the depth-$2$ case. 

We directly prove this theorem here.
\begin{proof}
We apply the Itô formula on $w_{t,+}^{2-p}$ and $w_{t,-}^{2-p}$ to get the following:
\begin{align*}
\dd [w_{t,+}^{2-p}] &= (2-p) w_{t,+}^{1-p} \odot \dd w_{t,+} + 2 (2-p)(1-p) \gamma L(\beta_t) w_{t,+}^{-p} \odot  w_{t,+}^{2p-2} \odot \diag (H) \\
&= -p(2-p) X^\top r(\beta_t) \dd t + 2 (2-p)(1-p) \gamma L(\beta_t)  w_{t,+}^{p-2} \odot \diag (H) \dd t + (2-p) \sqrt{\gamma L(\beta_t)} X^\top \dd B_t \\
&= -X^\top \dd A_t + C^+_t \dd t,
\end{align*}
where $\dd A_t := -p(p-2) r(\beta_t) \dd t + 2 (p-2) \sqrt{\gamma L(\beta_t)} \dd B_t$ and $C^+_t := 2 (p-2)(p-1) \gamma L(\beta_t)  w_{t,+}^{p-2} \odot \diag (H) $. Similarly, with explicit notations, we have that:
\begin{align*}
\dd [w_{t,-}^{2-p}] =   X^\top \dd A_t + C^-_t \dd t.
\end{align*}
Hence, 
\begin{align*}
w_{t,+}^{p} &=  \left[\alpha^{2-p} - X^\top \int_0^t \dd A_s + \int_0^t C^+_s \dd s\right]^{\frac{p}{2-p}} \quad \textrm{and} \quad w_{t,-}^{p} =  \left[\alpha^{2-p} + X^\top \int_0^t \dd A_s + \int_0^t C^-_s \dd s\right]^{\frac{p}{2-p}}.
\end{align*}
And finally,
\begin{align*}
\beta_{t} = w_{t,+}^{p} - w_{t,-}^{p} =   \left[\alpha^{2-p} + \int_0^t C^+_s \dd s - X^\top \int_0^t \dd A_s \right]^{\frac{p}{2-p}} - \left[\alpha^{2-p} + \int_0^t C^-_s \dd s + X^\top \int_0^t \dd A_s \right]^{\frac{p}{2-p}}.
\end{align*}
Defining $\alpha_{\textrm{eff},\pm}^{2-p} = \alpha^{2-p} + \int_0^\infty C^\pm_s \dd s$  and $\nu_\infty = \int_0^\infty \dd A_s$, if all quantities have limits when $t \to \infty$ we have that $\beta_\infty = h_{\alpha,p, \pm}(X^\top \nu_\infty)$, where $h_{\alpha,p, \pm}(z) = (\alpha_{\textrm{eff},+}^{2-p} - z)^{\frac{p}{2-p}} - (\alpha_{\textrm{eff},-}^{2-p} + z)^{\frac{p}{2-p}}$. Inverting this function and integrating gives the theorem with the standard KKT argument~\citep[see][ under Theorem $1$ page $4$]{woodworth2020kernel}.
\end{proof}
\label{app:extensions}


\section{Technical lemmas}

In this section, we state and prove technical lemmas which we use to prove our main results. 
\begin{lemma}
\label{app:lemma:martingale_result}
For any interpolator $\beta^*$, 
$S_t = \int_0^t  \sqrt{ \gamma L(\beta_s)} \langle \tX^{\top}   \dd B_s, \beta_s - \beta^* \rangle $ is a square-integrable martingale with a.s. continuous paths. And for any $a, b \geq 0$:
\begin{align*}
    P( \forall t \geq 0, \vert S_t \vert \leq a + 2 b \gamma \lambda_{\max} \int_0^t  L(\beta_s)   (\Vert \beta_s \Vert_1^2 +  \Vert \beta^{*} \Vert_1^2) \dd s) &\geq 1 - 2 \exp( - 2 a b)\\
&= 1 - p,
\end{align*}
where $p = 2 \exp( - 2 a b )$.

\end{lemma}

\begin{proof}

Since $(S_t)_{t \geq 0}$ is a is a locally square-integrable martingale with a.s. continuous paths, 
\cite[][Corollary 11]{10.1214/18-PS321} gives that
$$P (\exists t \in(0, \infty): S_{t} \geq a+b\langle S\rangle_{t} ) \leq \exp \{-2 a b\}).$$

We now compute the quadratic variation $ \langle S\rangle_{t} $.  Notice that 
$\langle \tX^{\top} \dd B_t, \beta_t - \beta^{*} \rangle = \sum_{k=1}^n [\tX (\beta_t - \beta^{*})]_k \dd B_t^k$,
hence the quadratic variation of $S_t$ equals:
\begin{align*}
\langle S \rangle_t &= \gamma \int_0^t L(\beta_s) \sum_{k=1}^n [\tX (\beta_t - \beta^{*})]_k^2 \dd s \\
&= \gamma \int_0^t L(\beta_s) \Vert  \tX (\beta_s - \beta^{*}) \Vert^2 \dd s \\
&= 4 \gamma \int_0^t L(\beta_s)^2 \dd s.
\end{align*}
Furthermore, since:
\begin{align*}
   4 \int_0^t  L(\beta_s)^2 \dd s &=    \int_0^t  L(\beta_s)  (\beta_s - \beta^{*})^T \tX^{\top} \tX (\beta_s - \beta^{*}) \dd s \\
    &\leq   \lambda_{\max} \int_0^t  L(\beta_s)   \Vert \beta_s - \beta^{*} \Vert_2^2 \dd s \\
    &\leq  2 \lambda_{\max} \int_0^t  L(\beta_s)   (\Vert \beta_s \Vert_2^2 +  \Vert \beta^{*} \Vert_2^2) \dd s \\
    &\leq  2 \lambda_{\max} \int_0^t  L(\beta_s)   (\Vert \beta_s \Vert_1^2 +  \Vert \beta^{*} \Vert_1^2) \dd s,
    \end{align*}
we obtain that: 

\begin{align*}
    \langle S \rangle_t &\leq     2 \gamma \lambda_{\max} \int_0^t  L(\beta_s)   (\Vert \beta_s \Vert_1^2 +  \Vert \beta^{*} \Vert_1^2) \dd s,
     \end{align*}
and:

\begin{align*}
P( \exists t \geq 0, \vert S_t \vert \geq a & + 2 b \gamma \lambda_{\max} \int_0^t  L(\beta_s)   (\Vert \beta_s \Vert_2^2 +  \Vert \beta^{*} \Vert_1^2) \dd s) \\
    &\leq P( \exists t \geq 0, \vert S_t \vert \geq a +  b \langle S \rangle_t)  \\
    &\leq  2 \exp (- 2 a b) .      
\end{align*}

\end{proof}

\begin{lemma}
\label{app:lemma:F:A_B_log_bound}
Let $A, B > 0$ such that $\frac{A}{B} + \ln(B) \geq 2$. 
Assume that $x \leq A + B \ln x$, then $$x \leq \frac{5}{2} (A + B \ln(B)).$$
\end{lemma}

\begin{proof}
$x \leq A + B \ln x$ is equivalent to $x \leq \exp(-\frac{A}{B}) \exp( \frac{x}{B})$. Standard analysis on the 
Lambert $W$ function shows that this leads 
to $x \leq -B \ W_{-1} ( -\frac{1}{B} \exp(- \frac{A}{B}))$, 
where $W_{-1}$ is the lower branch \footnote{see 
\url{https://en.wikipedia.org/wiki/Lambert_W_function}
for more details}. For $- \frac{1}{e} \leq z \leq 0$, the branch $W_{-1}$ can be lower bounded as:
$W_{-1}(z) \geq - \sqrt{-2 (1 + \ln(-z))} + \ln(-z)$ (see Theorem 1 of \cite{chatzigeorgiou2013bounds}). 
Since $\ln(-z) = \ln( \frac{1}{B} \exp(- \frac{A}{B})  ) = - ( \frac{A}{B} + \ln ( B ) )$:
\begin{align*}
x &\leq B ( \sqrt{2(- 1 + \frac{A}{B} + \ln(B)) } + \frac{A}{B} + \ln(B)   ) \\
&\leq B ( \sqrt{2} (- 1 + \frac{A}{B} + \ln(B))  + \frac{A}{B} + \ln(B)   ) \\
&\leq (\sqrt{2} + 1) B ( \frac{A}{B} + \ln(B)   ) \\
&\leq \frac{5}{2} (A + B \ln(B)   ).
\end{align*}
This concludes the proof of the Lemma.
\end{proof}

\begin{lemma}
\label{app:lemma:LoulouJeTeKiffe}
For any $ \alpha > 0$ and  $\beta \in\R$, we have the following inequality: $$ \phi_\alpha(\beta) - \phi_\alpha(0) \geq \frac{1}{4}  \max \Big \{ 0, \vert \beta  \vert \ln \frac{\vert \beta \vert}{2 \alpha^2} \Big \}.$$
\end{lemma}
\begin{proof}
Let us fix $\alpha \in \R$. First notice that by parity in $\beta$ of the functions involved, and as the inequality holds in $\beta=0$, we can suppose that $\beta>0$ and define $$f(\beta):= \phi_\alpha(\beta) - \phi_\alpha(0) = \frac{1}{4} \left[\beta \arcsinh \left(\frac{\beta}{2\alpha^2}\right) - \sqrt{\beta^2 + 4 \alpha^4} + 2 \alpha^2\right].$$
Trivially, $f'(\beta)= \frac{1}{4}\arcsinh \left(\frac{\beta}{2\alpha^2}\right)>0$. Hence, it increases on $\R_+$ and as $f(0) = 0$, $f$ is always positive. This show the inequality for the left term of the $\max$.

For the other term of the $\max$, let us define $g(\beta):= \frac{1}{4} \beta \ln \frac{ \beta }{2 \alpha^2}$, we have that $$4[f'(\beta) - g'(\beta)] =  \arcsinh \left(\frac{\beta}{2\alpha^2}\right)  - \ln \left(\frac{\beta}{2\alpha^2}\right) + 1 = \ln \left(1 + \sqrt{1 + \frac{4 \alpha^4}{\beta^2}}\right) + 1>0.$$
Hence, $f - g$ increases and as $f(0)  - g(0) = 0$, we have  that $f > g$ which concludes the proof.
\end{proof}


\label{app:technical_lemmas}

\end{document}